\documentclass[sigconf, screen]{acmart}
\AtBeginDocument{%
  }
\renewcommand\footnotetextcopyrightpermission[1]{}
\setcopyright{acmlicensed}
\copyrightyear{2018}
\acmYear{2018}
\acmDOI{XXXXXXX.XXXXXXX}
\acmConference[Conference acronym 'XX]{Make sure to enter the correct
  conference title from your rights confirmation email}{June 03--05,
  2018}{Woodstock, NY}
\acmISBN{978-1-4503-XXXX-X/2018/06}

\acmSubmissionID{963}

\usepackage{graphicx}
\usepackage{multirow}
\usepackage{colortbl}
\usepackage{makecell}
\usepackage{caption}
\usepackage{subcaption}
\usepackage{amsmath}
\usepackage{algorithm}
\usepackage{algorithmic}
\usepackage{color}
\usepackage{bm}
\usepackage{amsfonts}
\usepackage{arydshln}
\usepackage{pifont}
\usepackage{dashbox}
\usepackage{xspace}
\usepackage{siunitx}  
\usepackage{booktabs} 
\usepackage{array}    

\newcommand{\softmax}{\mathrm{softmax}}
\newcommand{\sigmoid}{\mathrm{sigmoid}}
\newcommand{\cmark}{\ding{51}}
\newcommand{\xmark}{\ding{55}}
\newcommand{\ours}{{\bf X-Agent}\xspace}
\newcommand{\hlrow}{\rowcolor{black!6}}

\begin{document}
\begin{sloppypar}

\title{Novel Category Discovery with X-Agent Attention for Open-Vocabulary Semantic Segmentation}


\author{Jiahao Li, Yang Lu, Yachao Zhang, Fangyong Wang, Yuan Xie$^*$, and, Yanyun Qu$^*$}\thanks{*Corresponding author. Code: \href{https://github.com/liblacklucy/X-Agent}{https://github.com/liblacklucy/X-Agent}}
\begin{abstract}
Open-vocabulary semantic segmentation (OVSS) conducts pixel-level classification via text-driven alignment, where the domain discrepancy between base category training and open-vocabulary inference poses challenges in discriminative modeling of latent unseen category. To address this challenge, existing vision-language model (VLM)-based approaches demonstrate commendable performance through pre-trained multi-modal representations. However, the fundamental mechanisms of latent semantic comprehension remain underexplored, making the bottleneck for OVSS. In this work, we initiate a probing experiment to explore distribution patterns and dynamics of latent semantics in VLMs under inductive learning paradigms. Building on these insights, we propose X-Agent, an innovative OVSS framework employing latent semantic-aware ``agent'' to orchestrate cross-modal attention mechanisms, simultaneously optimizing latent semantic dynamic and amplifying its perceptibility. Extensive benchmark evaluations demonstrate that X-Agent achieves state-of-the-art performance while effectively enhancing the latent semantic saliency.
\end{abstract}






\maketitle

\section{Introduction}

Open-vocabulary semantic segmentation~\cite{he2023primitive,zhao2017open,li2020consistent,shen2021conterfactual} (OVSS) seeks to assign dynamically variable open-category labels to all image pixels through textual prompt-driven multi-modal semantic alignment. The fundamental challenge lies in enabling zero-shot visual-semantic correspondence guided by textual descriptions of novel categories during inference, rendering the discriminative modeling of latent unseen category the primary performance bottleneck for OVSS. Existing approaches~\cite{zhou2022extract,liu2023delving,qin2023freeseg} primarily exploit the cross-modal alignment capabilities inherent in vision-language models~\cite{radford2021learning, jia2021scaling, singh2022flava} (VLMs) to circumvent this bottleneck while mitigating the intrinsic limitations of zero-shot generalization in inductive learning. Although these techniques (e.g., feature adaptation~\cite{zhou2023zegclip,kwon2023probabilistic} and knowledge distillation~\cite{ding2022decoupling,xu2022simple,yu2023zero,liang2023open}) demonstrate substantial performance improvements through VLM integration, their exploration of latent unseen semantics—especially concerning effective generalization mechanisms within inductive learning constraints—remains limited. This results in underdeveloped capabilities for discovering novel categories in VLM.

\begin{figure}[t]
  \includegraphics[width=\linewidth]{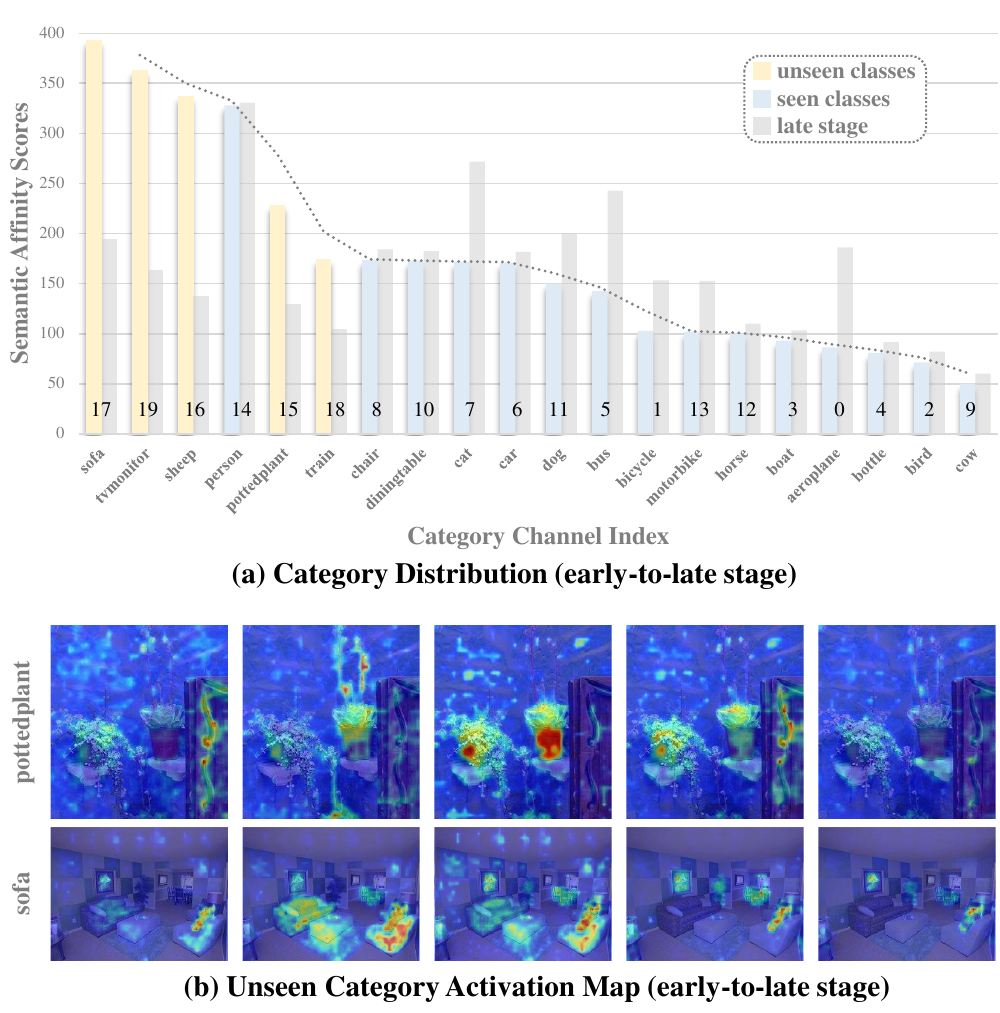}
  \caption{Our probing experiment results reveal that (a) latent semantic are predominantly distributed within category channels exhibiting high semantic affinity scores; (b) latent semantic demonstrate a phenomenon of gradual emergence followed by diminishing presence during training.}
  \Description{}
  \label{fig:motivation}
\end{figure}
We hypothesize this limitation originates in the inductive learning paradigm, whereby VLM-derived latent semantics become eclipsed by task-specific supervisory signals during training, thereby inducing progressive erosion of the model's discriminative reliability towards these latent semantics. To this end, we initiate a probing experiment to explore two fundamental problems: (1) the visual-semantic distribution patterns of latent semantic representations within VLM; and (2) the distribution dynamic of such representations under inductive training. Specifically, we perform linear probing on CLIP, adhering to the zero-shot setting where category spaces are partitioned into mutually exclusive seen and unseen sets. To analyze the distribution and dynamics of latent semantics, we carefully select 5,000 images (with predominant unseen category representation). As shown in Figure~\ref{fig:motivation}, our findings reveal that: (a) During initial training iterations, CLIP maintains strong discriminative capacity for unseen categories, attributable to its robust pre-trained representations as evidenced by high semantic affinity scores. (b) Inductive training phases induce gradual semantic degradation of unseen category discriminability, ultimately collapsing to random baseline performance, as verified through diminishing activation magnitudes (e.g., pottedplant and sofa). These empirical results substantiate that VLM-based approaches systematically accumulate task-specific inductive biases during training, thereby overwriting the foundational cross-modal knowledge inherent in pre-trained VLMs.

To preserve latent semantics in VLMs, maintaining their discriminative integrity presents a crucial research objective. This motivates our X-Agent framework, which implements three dedicated components (agent selection, agent pooling, and agent attention modules) specifically engineered for persistent latent semantic retention in OVSS tasks. Our core idea lies in 1) constructing a dedicated ``agent'' responsible for latent semantic perception, and then 2) integrating it into the cross-modal attention mechanism to amplify the latent semantic saliency. Therefore, X-Agent addresses two critical challenges: 1) architectural design of latent-semantic-aware ``agent'', and 2) topological-consistent integration of agent-mediated attention mechanisms. For the first challenge, the agent selection component implements latent semantic tokens based on the probing experiment while the agent pooling component aggregates multi-modal global receptive fields. For the second challenge, agent attention component utilizes the ``agent'' as parametric intermediaries in cross-modal attention mechanisms, enforcing topological consistency through constrained cross-modal feature interactions. Note that our ``agent'' fundamentally differs from LLM agent in architectural design. Due to sharing functional similarities as task-oriented intermediate architectures (e.g., our ``agent'' specializes in the perception and enhancement of latent semantic), we adopt this nomenclature. Our principal contributions are as follows:
\begin{itemize}
\item We empirically investigate the distribution patterns and dynamics of latent semantics in VLM under inductive learning paradigms.
\item We propose X-Agent, a novel OVSS method that aims to enhance latent semantic saliency through agent-mediated cross-modal attention mechanisms.
\item We achieve superior performance on standard benchmarks, while effectively enhancing the latent semantic saliency.
\end{itemize}

\begin{figure*}[th]
  \includegraphics[width=\textwidth]{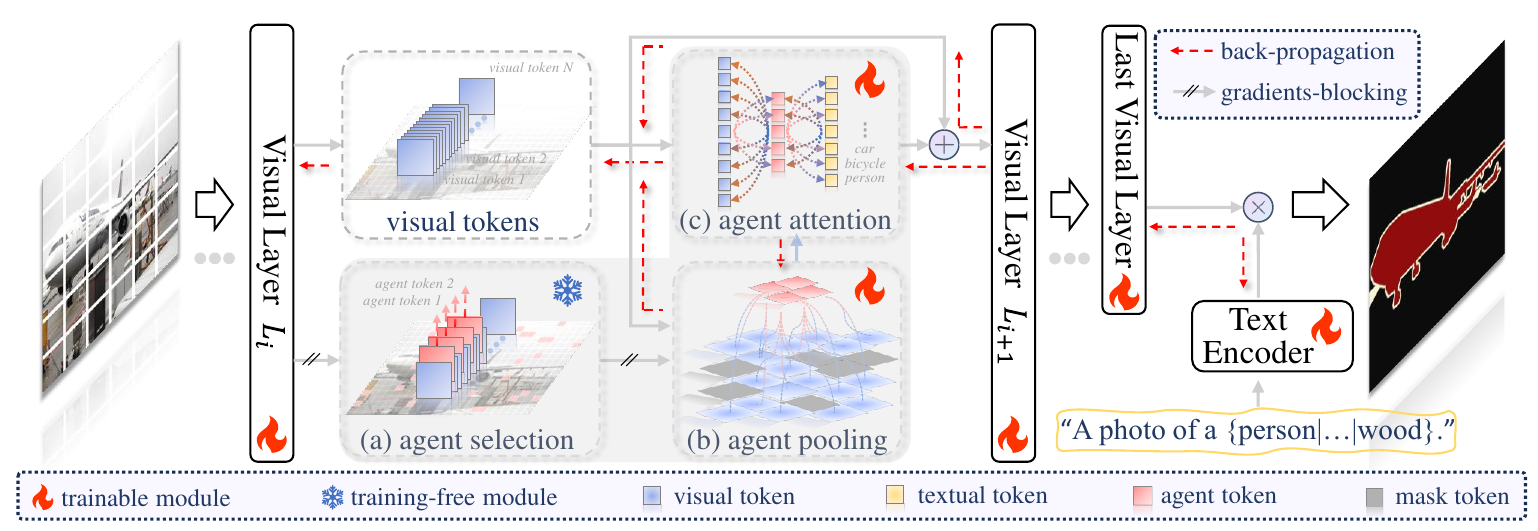}
  \caption{Overview of X-Agent. Our X-Agent framework operates directly on the cross-layer visual embeddings of VLM, e.g., CLIP, to strengthen latent semantic perception capability. The architecture consists of three synergistic components: (1) agent selection module, designed to excavate latent semantic tokens from cross-layer visual embeddings and select these tokens as ``agent'' for latent semantic perception; (2) agent pooling module, aimed at broadening the multi-modal receptive field of the ``agent'' dedicated to latent semantic perception; (3) agent attention module, which leverages the ``agent'' to modulate visual attention mechanism to enhance latent semantic discriminability.}
  \Description{}
  \label{fig:overview}
\end{figure*}

\section{Related Works}

\noindent \textbf{Open-Vocabulary Semantic Segmentation.} OVSS methodologies predominantly dichotomize into two principal paradigms: 1) generative approaches~\cite{gu2020context,bucher2019zero,cheng2021sign,wu2023diffumask,wang2025diffusion} under transductive learning frameworks, and 2) discriminative methods~\cite{ding2022decoupling,zhou2022extract,zhou2023zegclip,xian2019semantic,shi2025llmformer} following inductive learning principles. Generative methods fall under transductive methodologies, typically requiring a prior knowledge about unseen categories in open-world scenarios. These techniques~\cite{xian2019semantic,bucher2019zero,pastore2021closer,gu2020context} synthesize virtual unseen semantic embeddings by amalgamating visual embeddings with textual semantic embeddings from existing priors. Discriminative approaches constitute inductive methodologies that infer unseen semantics through learned categorical knowledge during training, thereby eliminating the need for prior knowledge about novel categories. State-of-the-art implementations predominantly employ either knowledge distillation~\cite{ding2022decoupling,xu2022simple,yu2023zero,liang2023open} or feature adaptation strategies~\cite{zhou2023zegclip,kwon2023probabilistic,xu2023side,xie2023sed,cho2023cat,li2024relationship}. Knowledge distillation methods typically combine VLM-derived image-level semantic discriminability with mask-aware segmentation networks to achieve open-vocabulary segmentation, while feature adaptation approaches directly fine-tune VLMs as backbone networks to convert image-level classification capabilities into pixel-level discriminative power. Despite remarkable progress, current methodologies exhibit insufficient exploration of latent semantic perception mechanisms. 

\noindent \textbf{Generalized Category Discovery.} Generalized category discovery constitutes a specialized form of out-of-distribution (OOD) problems~\cite{yang2024generalized}. Current research in Novel category discovery predominantly focuses on two paradigms: novel category discovery~\cite{han2019learning} and open-set recognition~\cite{Scheirer_2013_TPAMI}. The former paradigm addresses the classification of unseen category instances through knowledge transfer from observed categories, typically implemented via multi-stage training frameworks~\cite{Hsu18_L2C,Hsu19_MCL,han21autonovel,han20automatically}. State-of-the-art methods employ consistency regularization~\cite{ncd-ncl,wta} and pseudo-labeling mechanisms~\cite{openmix,uno} to mitigate domain-specific semantic distribution shifts. The latter paradigm serves as a generalized form of novel category discovery where test samples may contain both known and novel categories~\cite{yang2024generalized,Scheirer_2013_TPAMI}. Some prior works~\cite{masana2018metric,shu2020p,liu2020few,chen2020learning} explore prototype-based contrastive learning with rank-ordered distance metrics for novel category discrimination; alternative approaches~\cite{zhang2016sparse,liu2017incremental,sun2020conditional,shao2020open} investigate reconstruction-based modeling of novel class distributions. In this work, we focus on how intrinsic semantic representations in VLMs preserve generalization capacity without semantic dissipation under inductive learning, distinct from the semantic shift problems addressed by the above methodologies.

\section{Approach}

\subsection{Problem Definition} \label{Problem Definition}
Given an image ${\bm I}$ and a set of category-specific textual descriptions $\{{\bm T(i)}\}^{N_c}_{i=1}$, where ${\bm T(i)}$ denotes the textual descriptions of the $i$-th class and $N_c$ represents the total number of classes, the OVSS network aligns each pixel within the image ${\bm I}$ to the most semantically relevant textual description $T(i)$ from the set $\{{\bm T(i)}\}^{N_c}_{i=1}$, thereby assigning the corresponding class label $i$ to the pixel. Note that the total number of classes $N_c$ is dynamic at inference time. 

\subsection{Overview} \label{Overview}
Our objective is to enhance the generalization capability of OVSS networks in perceiving latent unseen semantics. To address this, we present X-Agent, a novel framework that discovers latent semantic embeddings in high-dimensional visual feature spaces, functioning these embeddings as the ``agent'' to adaptively enhance latent semantic saliency. As illustrated in Figure~\ref{fig:overview}, X-Agent consists of three key components: (a) {\bf agent selection}, which aims to localize latent semantic tokens and select them as a set of agent tokens; (b) {\bf agent pooling}, which aggregates multi-modal global-contextual knowledge for the agent tokens, establishing the ``agent'' to perceive latent semantic distribution; (c) {\bf agent attention}, which utilizes the agent tokens to intervene in the core attention mechanism in VLM to enhance latent semantic saliency.

Specifically, the OVSS network follows a multi-modal encoder-decoder architecture, where multi-modal encoder (e.g., CLIP) contains a visual encoder $\mathcal{V}$ and a text encoder $\mathcal{T}$. The image ${\bm I}$ and a set of category-specific textual descriptions $\{{\bm T(i)}\}^{N_c}_{i=1}$ are fed into the visual encoder $\mathcal{V}$ and text encoder $\mathcal{T}$ to produce the visual and the textual embeddings, respectively. These embeddings are subsequently passed to the decoder $\mathcal{D}$ to generate the segmentation results ${\bm O}$. This process can be formally expressed as:
\begin{equation}
  \bm{O} = \mathcal{D}(\mathcal{V}({\bm I}), \mathcal{T}(\{{\bm T(i)}\}^{N_c}_{i=1})).
\end{equation}
X-Agent $\mathcal{X}$ operates on intermediate layer embeddings of the visual encoder $\mathcal{V}$, taking the visual tokens ${\bm f}_v \in \mathbb{R}^{N\times d}$ from each visual layer as input, where $N$ and $d$ denote the number of the patch tokens and the embedding dimension, respectively, as follows:
\begin{equation}
  \bm{O} = \mathcal{D}((\mathcal{V} \circ \mathcal{X})({\bm I}), \mathcal{T}(\{{\bm T(i)}\}^{N_c}_{i=1})).
\end{equation}
In the following sections, we will elaborate on the details of the three proposed core components.

\subsection{Agent Selection} \label{Agent Selection}
\begin{figure}[t!]
  \centering
  \includegraphics[width=\linewidth]{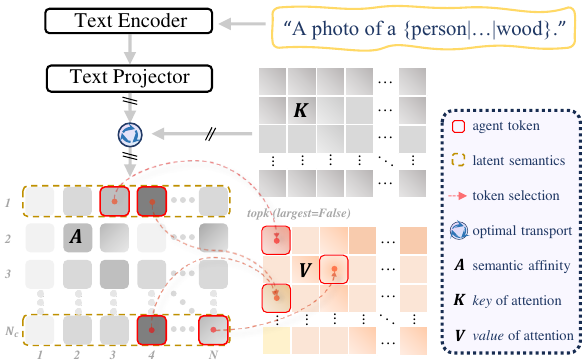}
  \caption{Agent Selection. The semantic affinity matrix is constructed through optimal transport between textual and visual representations, followed by the selection of agent tokens guided by a top-$k$ strategy.}
  \Description{}
  \label{fig:method1}
\end{figure}

Enhancing the OVSS network's ability to perceive latent semantics necessitates precise localization of latent unseen category tokens, referred to as agent tokens. Our agent selection component operates through two steps: (1) computing a cross-modal semantic affinity matrix through optimal transport-guided alignment, and (2) identifying latent unseen category tokens as agent tokens via top-$k$ selection based on the affinity matrix. As depicted in Figure~\ref{fig:method1}, we formulate the semantic affinity matrix using optimal transport between the \textit{key} matrix of the visual attention block and the textual tokens, with agent tokens selected from the visual attention \textit{value} matrix using a top-$k$ strategy.

Specifically, the category-specific textual descriptions $\{{\bm T(i)}\}^{N_c}_{i=1}$ (e.g., {``A photo of a car''}) are encoded through CLIP's text encoder and projected to the dimensions of the visual token ${\bm f}_v$ via a lightweight Transformer-based text projector, yielding the textual tokens ${\bm f}_t\in \mathbb{R}^{N_c\times d}$. Given the visual attention \textit{key} matrix ${\bm K}\in \mathbb{R}^{N\times d}$, the affinity matrix $\bm{A} \in \mathbb{R}^{N_c\times N}$ can initially be defined as the cosine similarity between the textual tokens ${\bm f}_t$ and the \textit{key} matrix ${\bm K}$.  However, empirical studies demonstrate that this approach fails to effectively capture fine-grained semantic relationships. To overcome this limitation, we employ the Sinkhorn algorithm~\cite{cuturi2013sinkhorn} to compute the optimal transport plan ${\bm P^*\in \mathbb{R}^{N_c\times N}}$ between the textual tokens ${\bm f}_t$ and the \textit{key} matrix ${\bm K}$, thereby refining the affinity matrix $\bm{A}$. According to the Sinkhorn algorithm, the empirical distributions $\bm{\mu}$ and $\bm{\nu}$ for $\bm{f}_t$ and $\bm{K}$ are defined as:
\begin{equation}
  \bm{\mu} = \sum^{N_c}_{i=1} \mu_i \delta_{\bm{f}^i_t}, \quad \bm{\nu} = \sum^N_{i=1} \nu_i \delta_{\bm{K}^\top},
\end{equation}
where $\delta_{\bm{f}_t}$ and $\delta_{\bm{K}^\top}$ denote Dirac functions centered on $\bm{f}_t$ and $\bm{K}^\top$, respectively. The optimal transport plan $\bm{P}^*$ is then calculated as:
\begin{equation}
  \bm{P}^* = \operatorname{diag}({\bm a}^j) \exp(-{\bm C}/\epsilon) \operatorname{diag}({\bm b}^j),
\end{equation}
where $j$ is the iteration, ${\bm a}^j = \bm{\mu} / \exp(-{\bm C}/\epsilon) {\bm b}^{j-1}$ and ${\bm b}^j = \bm{\nu} / \exp(-{\bm C}/\epsilon) {\bm a}^j$, with $\bm{b}^0 = \mathbf{1}$, $\epsilon > 0$ is the regularization scalar, and $\bm{C}$ is the cost matrix measuring the distance between the textual tokens ${\bm f}_t$ and the \textit{key} matrix ${\bm K}$. Thus, we set the cost matrix $\bm{C}$ as:
\begin{equation}
  \bm{C} = \mathbf{1} - \frac{\bm{f}_t}{\operatorname{max}(\|\bm{f}_t\|_2)} \cdot  \frac{\bm{K}^\top}{\operatorname{max}(\|\bm{K}^\top\|_2)},
  \label{eq:C}
\end{equation}
where $\|\cdot\|_2$ denotes the $L_2$ normalization. Finally, the refining semantic affinity matrix ${\bm A}$ can be defined as:
\begin{equation}
  \bm{A} = \sigmoid( \bm{P}^* \odot (\frac{\bm{f}_t}{\|\bm{f}_t\|_2} \cdot \frac{\bm{K}^\top}{\|\bm{K}^\top\|_2})),
\end{equation}
where $\odot$ denotes the element-wise product. To discover latent semantic tokens, based on our probing experiment (Figure~\ref{fig:motivation} (a)), our analysis reveals that the latent semantics predominantly reside in the top-$k$ category channels of the affinity matrix ${\bm A}$. From these $k$ channels, we select the latent-related top-$q$ tokens and, guided by the indices of these tokens, select corresponding tokens from the visual attention \textit{value} matrix ${\bm V}$ as agent tokens ${\bm f}_x \in \mathbb{R}^{(k\cdot q)\times d}$. These top-$k$ category channels ${\bm A}^*\in \mathbb{R}^{k\times N}$ can be defined as:
\begin{align}
  \operatorname{idx}_1 &= \operatorname{top-}k(\operatorname{mean}(\bm{A},dim=-1), k), \\
  {\bm A}^* &= \operatorname{gather}({\bm A}, \operatorname{idx}_1),
\end{align}
where $\operatorname{top-}k$, $\operatorname{mean}$, and $\operatorname{gather}$ operations leverage PyTorch's built-in operators. The token selection process can be defined as:
\begin{align}
  \operatorname{idx}_2 &= \operatorname{top-}k(\bm{A}^*, q, \text{largest=False}), \\
  {\bm f}_x &= \operatorname{gather}({\bm V}, \operatorname{idx}_1 \& \operatorname{idx}_2).
\end{align}
Considering overlapping indices during the token selection process, we introduce mask tokens to replace duplicate ones within the agent tokens, thereby avoiding redundant selection of the same token from the \textit{value} matrix ${\bm V}$. This mechanism effectively eliminates duplicate token sampling from the \textit{value} matrix , preventing semantic bias introduced by repetitive feature aggregation.

\subsection{Agent Pooling} \label{Agent Pooling}
\begin{figure}[t]
  \centering
  \includegraphics[width=\linewidth]{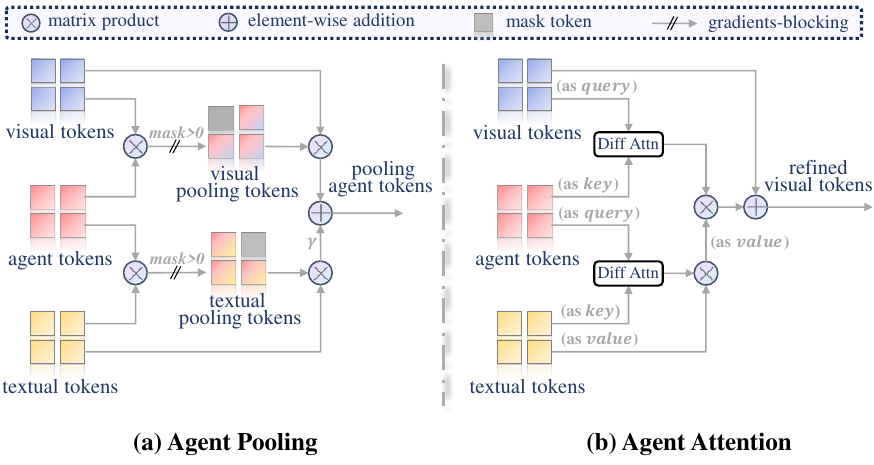}
  \caption{(a) Agent Pooling. The agent tokens are fed into a mask-guided parallel dual-branch pooling architecture to fuse cross-modal contexts while enlarging effective receptive fields. (b) Agent Attention. Employing cascaded differential attention, the agent tokens first operate as \textit{query} to extract latent semantics from textual tokens, then act as \textit{key} to guide visual token refinement.}
  \Description{}
  \label{fig:method2_3}
\end{figure}

To strengthen the semantic perception capacity of visual representations, an intuitive approach lies in infusing semantic knowledge through cross-attention mechanisms between textual and visual embeddings. However, this paradigm suffers from homogenized category contributions that hinder novel category discovery. We address this by introducing a mediator ``agent'' operating in latent semantic space between textual and visual embeddings, implemented as the agent tokens. Serving as an adaptive bridge, these agent tokens require comprehensive multi-modal contextual awareness to establish fine-grained alignment between modalities. As illustrated in Figure~\ref{fig:method2_3} (a), our agent pooling component employs a mask-guided dual-branch parallel architecture that simultaneously handles: (1) vision-guided contextual aggregation, and (2) text-aware semantic aggregation. This dual-stream framework synergizes modality-specific inductive biases while expanding receptive fields through cross-modal feature fusion, endowing the agent tokens with enhanced multi-modal perception capabilities.

Specifically, given the visual tokens ${\bm f}_v$ and the agent tokens ${\bm f}_x$, the visual mask tokens ${\bm f}_{v_m}\in \mathbb{R}^{N\times (k\cdot q)}$ are computed through:
\begin{equation}
  \bm{f}_{v_m} = \frac{( {\bm f}_v \cdot {\bm f}^\top_x > 0 )}{\sum^{N-1}_{i=0}( {\bm f}_v \cdot {\bm f}^\top_x > 0 )_{i,j}}.
  \label{eq:mask}
\end{equation}
The visual pooling tokens ${\bm f}_{v_p}\in \mathbb{R}^{(k\cdot q)\times d}$ are subsequently obtained via:
\begin{equation}
  \bm{f}_{v_p} = \operatorname{proj}({\bm f}_{v_m}^\top \cdot {\bm f}_v),
  \label{eq:pooling}
\end{equation}
where $\operatorname{proj}$ denotes a single linear layer. Symmetrically, given the textual tokens ${\bm f}_t$, the textual pooling tokens ${\bm f}_{t_p}$ is derived using Equations~\ref{eq:mask} and ~\ref{eq:pooling}. The final agent tokens emerge through element-wise summation:
\begin{equation}
  \bm{f}_x = \bm{f}_{v_p} + \gamma\bm{f}_{t_p},
  \label{eq:agent}
\end{equation}
where $\gamma\in \mathbb{R}$ is a learnable scalar. To synchronize the learning dynamics, we re-parameterize the scalar $\lambda$ as:
\begin{equation}
    \label{eq:gamma}
    \gamma = ( \exp( \gamma_{v} ) - \exp( \gamma_{t}) ) + \gamma_{\text{init}},
\end{equation}
where $\gamma_{v}$ and $\gamma_{t}$ are learnable scalars, $\gamma_{\text{init}}\in (0,1)$ is a fixed initialization constant. Throughout subsequent formulations, all references to agent tokens strictly adhere to the definition in Equation~\ref{eq:agent}.

\subsection{Agent Attention} \label{Agent Attention}
While vision-language cross-attention mechanisms enable multi-modal feature interaction, they inherently suffer from semantic homogenization and inadequate latent semantic perception. Inspired by~\cite{han2024agent,ye2024differential}, we propose agent attention—a novel paradigm extending the conventional cross-attention triplet $(\bm Q, \bm K, \bm V)$ to a quadruplet formulation $(\bm Q, \bm X, \bm K, \bm V)$ with the agent tokens $\bm X$ and replacing standard attention with differential attention—as depicted in Figure~\ref{fig:method2_3}(b). These agent tokens act as latent semantic ``agent'', enabling differential attention computation through two cascaded attention blocks. The first block establishes semantic-aware \textit{value} matrix via agent-and-text attention, while the second performs agent-and-visual attention to effectively disentangling latent semantics in visual tokens.

Specifically, given the visual tokens ${\bm f}_v$ and the textual tokens ${\bm f}_t$, we instantiate the cross-attention triplet $(\bm Q, \bm K, \bm V)$ by designating ${\bm f}_v$ as \textit{query} matrix $\bm Q$ and ${\bm f}_t$ as \textit{key} and \textit{value} matrices $(\bm K, \bm V)$. The standard cross-attention operation computes:
\begin{equation}
    \label{eq:cross}
    \operatorname{CrossAttn}=(\softmax(\frac{\bm Q \bm K^{T}}{\sqrt{d}})){\bm V},\quad \bm Q={\bm f}_v, \quad \bm K=\bm V={\bm f}_t.
\end{equation}
In our agent attention, given the agent tokens ${\bm f}_x$ and the textual tokens ${\bm f}_t$, we refer to ${\bm f}_x$ as \textit{query} matrix $\bm Q$ and ${\bm f}_t$ as \textit{key} and \textit{value} matrices $(\bm K,\bm V)$. The first differential attention block operates on the agent-text interaction:
\begin{align}
    \bm V_x&=\operatorname{DiffAttn}(\bm Q,\bm K,\bm V), \quad \bm Q={\bm f}_x, \quad \bm K=\bm V={\bm f}_t,\\
    \operatorname{DiffAttn}&=(\softmax(\frac{\bm Q_1 \bm K^{T}_1}{\sqrt{d}}) - \lambda~\softmax(\frac{\bm Q_2 \bm K^{T}_2}{\sqrt{d}})){\bm V}^{\prime}, \\
    [\bm Q_1 ; \bm Q_2] &= \bm Q \bm W^Q ,\quad [\bm K_1 ; \bm K_2] = \bm K\bm  W^K ,\quad \bm V^{\prime} = \bm V \bm W^V,
\end{align}
where projection matrices $\bm W^Q, \bm W^K, \bm W^V \in \mathbb{R}^{d\times 2d}$ enable dimensional expansion and $\lambda\in \mathbb{R}$ is a learnable scalar. Subsequently, we refer to the visual tokens ${\bm f}_v$ and the agent tokens ${\bm f}_x$ as \textit{query} $\bm Q$ and \textit{key} $\bm K$, respectively. The second differential attention block processes visual-agent interaction:
\begin{equation}
    \Delta{\bm f}_v=\operatorname{DiffAttn}(\bm Q,\bm K,\bm V_x), \bm Q={\bm f}_v, \bm K={\bm f}_x.
\end{equation}
The complete agent attention mechanism is thus formulated as:
\begin{equation}
    \Delta{\bm f}_v=\operatorname{DiffAttn}({\bm f}_v, {\bm f}_x, \operatorname{DiffAttn}({\bm f}_x,{\bm f}_t,{\bm f}_t)),
\end{equation}
where ${\bm f}_v, {\bm f}_x$, and ${\bm f}_t$ denote $\bm Q,\bm X,$ and $(\bm K,\bm V)$ in the quadruplet $(\bm Q,\bm X,\bm K,\bm V)$, respectively. Finally, the refined visual tokens ${\bm f}^\prime_v$ are obtained via residual connection:
\begin{equation}
    {\bm f}^\prime_v = {\bm f}_v + \Delta{\bm f}_v.
\end{equation}
Given that the differential attention exhibit dual-function mechanisms combining adaptive noise suppression and context-sensitive attention refinement, through strategic deployment of agent tokens serving dual roles as \textit{query}/\textit{key} projections in cross-modal alignment processes, our agent attention component establish text-to-visual latent semantic propagation while preserving topological consistency.

\subsection{Loss Function} \label{Loss Function}
A fundamental challenge in infusing textual semantics into cross-layer visual representations stems from the dimensional incompatibility between linguistic and visual feature spaces. To bridge this gap, we introduce a learnable text projector (as shown in Figure~\ref{fig:method1}) comprising a lightweight Transformer layer for dimensional compatibility. However, this projection introduces potential distribution shift risks in the original textual embedding space. Therefore, we design a textual alignment loss to preserve semantic integrity.

Specifically, given the initial textual embeddings ${\bm f}_{t_{init}}\in \mathbb{R}^{N_c\times d^\prime}$ from the pretrained text encoder, we perform dimensional alignment through a single linear layer $\mathcal{\bm \phi}:\mathbb{R}^{d^\prime} \rightarrow \mathbb{R}^d$ to obtain a new textual embeddings ${\bm f}^\prime_{t}\in \mathbb{R}^{N_c\times d}$. To preserve semantic consistency during projection, we formulate a contrastive alignment loss with dual-directional feature normalization:
\begin{align}
    {\bm S}_1 &= (\frac{\bm{f}_t}{\|\bm{f}_t\|_2} \cdot \frac{\bm{f}^{\prime \top}_t}{\|\bm{f}^{\prime \top}\|_2})/\tau_1, \\
    {\bm S}_2 &= (\frac{\bm{f}^\prime_t}{\|\bm{f}^\prime_t\|_2} \cdot  \frac{\bm{f}^\top_t}{\|\bm{f}_t^\top\|_2})/\tau_2, \\
    \mathcal{L}_{\text{align}}&=\frac{1}{2}(\operatorname{CE}({\bm S}_1, \bm{I})+\operatorname{CE}({\bm S}_2, \bm{I})),
\end{align}
where $\bm{I}\in \mathbb{R}^{N_c\times N_c}$ denotes the identity matrix indicating feature correspondence, $\operatorname{CE}$ denotes the cross-entropy loss function, and $\tau_1,\tau_2$ are the learnable temperature parameter. The total optimization objective combines task-specific supervision:
\begin{equation}
    \mathcal{L} = \mathcal{L}_{\text{seg}}+\mathcal{L}_{\text{align}},
\end{equation}
where $\mathcal{L}_{\text{seg}}$ denotes the standard cross-entropy loss for open-vocabulary semantic segmentation.

\begin{table*}[!t]
\centering
\small
\caption{\textbf{Quantitative evaluation on standard benchmarks (unit:\%).} $\dagger$: Re-implementation trained on full COCO-Stuff.}
\label{tab:ov}
\resizebox{\textwidth}{!}{
    \begin{tabular}{c|cccc|cccccc}
    \toprule
    Model & VLM & Additional Backbone & Training Dataset & Additional Dataset & A-847 & PC-459 & A-150 & PC-59 & PAS-20 & PAS-21\\
    \midrule\midrule
    SPNet~\citep{xian2019semantic}  & - & ResNet-101 & PASCAL VOC & \xmark & - & - & - & 24.3 & 18.3 & - \\
    ZS3Net~\citep{bucher2019zero}  & - & ResNet-101 & PASCAL VOC & \xmark  & - & - & - & 19.4 & 38.3 & - \\
    LSeg~\citep{li2022language} & CLIP ViT-B/32 & ResNet-101 & PASCAL VOC-15  & \xmark & - & - & - & - & 47.4 & - \\
    LSeg+~\citep{ghiasi2022scaling} & ALIGN & ResNet-101 & COCO-Stuff & \xmark  & 2.5 & 5.2 & 13.0 & 36.0 & - & 59.0 \\
    ZegFormer~\citep{ding2022decoupling} & CLIP ViT-B/16 & ResNet-101 & COCO-Stuff-156 & \xmark   & 4.9 & 9.1 & 16.9 & 42.8 & 86.2 & 62.7 \\
    ZegFormer$\dagger$~\citep{ding2022decoupling} & CLIP ViT-B/16 & ResNet-101 & COCO-Stuff & \xmark & 5.6 & 10.4 & 18.0 & 45.5 & 89.5 &{65.5} \\
    ZSseg~\citep{xu2022simple} & CLIP ViT-B/16 & ResNet-101 & COCO-Stuff & \xmark   & 7.0 & - & 20.5 & 47.7 & 88.4 & - \\
    OpenSeg~\citep{ghiasi2022scaling} & ALIGN & ResNet-101 & COCO Panoptic & \cmark  &  4.4 & 7.9 & 17.5 & 40.1 & - & 63.8 \\
    OVSeg~\citep{liang2022open} & CLIP ViT-B/16 & ResNet-101c & COCO-Stuff & \cmark  &  7.1 & 11.0 & 24.8 & 53.3 & 92.6 & - \\
    ZegCLIP~\citep{zhou2023zegclip} & CLIP ViT-B/16 & - & COCO-Stuff-156 & \xmark  & -&-&-&41.2& 93.6 & - \\
    SAN~\citep{xu2023side} & CLIP ViT-B/16 & - & COCO-Stuff & \xmark  &{10.1} & {12.6} & {27.5} & {53.8} &{94.0} & - \\
    SED~\cite{xie2023sed}   & CLIP ConvNeXt-B  & - & COCO-Stuff    & \xmark    &{11.4} &18.6       & 31.6  &{57.3} &{94.4}  & -  \\
    CAT-Seg~\cite{cho2023cat} & CLIP ViT-B/16 & - & COCO-Stuff & \xmark & {\bf 12.0} &\underline{19.0} &\underline{31.8} &\underline{57.5} &{94.6} &\underline{77.3} \\
    RPN~\cite{li2024relationship} & CLIP ViT-B/16 & - & COCO-Stuff & \xmark &{11.4} &{17.3} &{31.5}  & 57.1   &\bf 95.2   \\
    \hlrow 
    \ours (ours) & CLIP ViT-B/16 & - & COCO-Stuff & \xmark &\underline{11.9} &{\bf 19.2} &{\bf 32.1} &{\bf 57.7} &\underline{95.1} &{\bf 77.4} \\
    \midrule
    LSeg~\citep{li2022language} & CLIP ViT-B/32 & ViT-L/16 & PASCAL VOC-15 & \xmark & - & - & - & - & 52.3 & - \\
    OpenSeg~\citep{ghiasi2022scaling} & ALIGN & Eff-B7 & COCO Panoptic & \cmark & 8.1 & 11.5 & 26.4 & 44.8 & - &{70.2}\\
    OVSeg~\citep{liang2022open} & CLIP ViT-L/14 & Swin-B & COCO-Stuff & \cmark & 9.0 & 12.4 & 29.6 & 55.7 & 94.5 & - \\
    SAN~\citep{xu2023side} & CLIP ViT-L/14 & - & COCO-Stuff & \xmark &{12.4} &{15.7} &{32.1} &{57.7} &{94.6} & - \\
    ODISE~\citep{xu2023open} & CLIP ViT-L/14 & Stable Diffusion & COCO-Stuff & \xmark & 11.1 & 14.5 & 29.9 & 57.3 & - & - \\
    SED~\cite{xie2023sed}    & CLIP ConvNeXt-L  & - & COCO-Stuff    & \xmark    &13.9 &22.6       &{35.2}  & 60.6 & 96.1 & -   \\
    CAT-Seg~\cite{cho2023cat} & CLIP ViT-L/14 & - & COCO-Stuff & \xmark &{\bf 16.0} &\underline{23.8} &\underline{37.9} &\underline{63.3} &\underline{97.0} &\underline{82.5} \\
    RPN~\cite{li2024relationship} & CLIP ViT-L/14 & - & COCO-Stuff & \xmark &14.9 &{22.1} &36.4  &{61.9}   &{96.6}   \\
    \hlrow 
    \ours (ours) & CLIP ViT-L/14 & - & COCO-Stuff & \xmark &{\bf 16.0} &{\bf 24.2} &{\bf 38.2} &{\bf 63.7} &{\bf 97.6} &{\bf 82.7} \\
    \bottomrule
    \end{tabular}
}
\end{table*}
\begin{table}[t]
\tiny
\centering
\caption{{\bf Performance comparison in the zero-shot setting (unit:\%).} Here, the best results are shown in bold and the second-best results are underlined.}
\label{tab:zs}
\resizebox{1\linewidth}{!}{
    \begin{tabular}{c|cccc}
    \hline
    {Modal} & {pAcc}  & {mIoU\(_\text{s}\)}  & {mIoU\(_\text{u}\)}  & {hIoU}   \\
    \hline
    ZegFormer~\cite{ding2022decoupling}      &-       &36.6          &33.2          &34.8        \\
    ZegFormer+MAFT~\cite{jiao2023learning}   &-       &36.4          &40.1          &38.1        \\
    ZSSeg~\cite{xu2022simple}                &60.3    &39.3          &36.3          &37.8        \\
    ZSSeg +MAFT~\cite{jiao2023learning}      &-    &36.1          &35.9          &36.0           \\
    ZegCLIP~\cite{zhou2023zegclip}           &{62.0}    &40.2          &41.4          &40.8      \\
    FreeSeg~\cite{qin2023freeseg}            &-    &\underline{42.4}         &{42.2}         &\underline{42.3}       \\
    {RPN~\cite{li2024relationship}}       &\underline{64.4}       & {40.8}         &\underline{42.8}    & {41.8} \\ 
    \hlrow
    {\ours (ours)}      &{\bf 65.7}       & {\bf 43.3}         &{\bf 42.9}          & {\bf 43.1}         \\ 
    \hline
    \end{tabular}
}
\end{table}

\section{Experiments}

\subsection{Experiment Settings} \label{Experiment Settings}
\noindent \textbf{Datasets and Metric.} Following prior works, we train on the COCO-Stuff dataset~\cite{caesar2018cvpr} and evaluate zero-shot generalization on three benchmarks: ADE20K~\cite{zhou2019semantic}, Pascal VOC2012~\cite{pascal-voc-2012}, and Pascal Context~\cite{mottaghi_cvpr14}. ADE20K contains 25,000 training and 2,000 validation images, with two evaluation configurations: A-150 (150 classes) and A-847~\cite{ding2022decoupling} (847 classes). Pascal VOC2012 includes 10,582 training and 1,449 validation images, evaluated under PAS-20 (20 object classes) and PAS-21 (21 classes, including the ``background'' category). Pascal Context comprises 10,100 images split into 4,996 training and 5,104 validation samples, with two configurations: PC-59 (59 classes) and PC-459 (459 classes). We evaluate all experiments using the mean Intersection-over-Union (mIoU) metric.

\noindent \textbf{Implementation.} All experiments are implemented using PyTorch~\cite{paszke2019pytorch} and Detectron2~\cite{wu2019detectron2} on four NVIDIA RTX 3090 GPUs. We adopt CLIP as the vision-language encoders $\mathcal{V}$ and $\mathcal{T}$, fine-tuning only the \textit{query} and \textit{value} linear layers within their attention modules. The decoder $\mathcal{D}$ employs a convolutional layer with bilinear upsampling, consistent with the design in CAT-Seg~\cite{cho2023cat}. To enhance parameter efficiency, we employ cross-layer weight-sharing for agent attention component, set $k=10$ and $q=4$ (i.e., all 40 agent tokens), and $\gamma_{init}=0.1$. Training configurations include: a batch size of 4, input resolution of $384\times384$, and 80k training iterations. We use the AdamW optimizer with an initial learning rate of $2\times10^{-4}$ for the decoder $\mathcal{D}$ and $2\times10^{-6}$ for CLIP, coupled with a weight decay of $10^{-4}$.

\subsection{System Level Comparison}
\begin{figure*}[t]
  \centering
  \includegraphics[width=\linewidth]{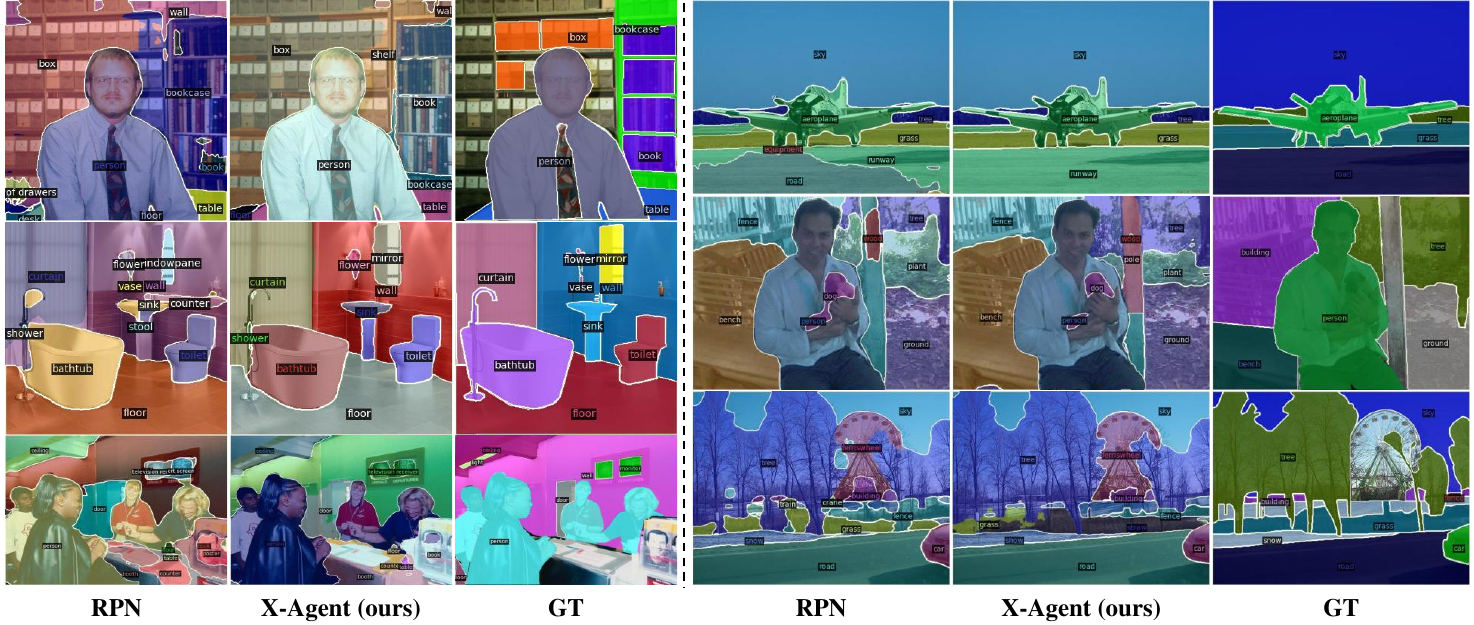}
  \caption{Qualitative comparison to RPN~\cite{li2024relationship}. Visualization results for the A-150 and PC-459 benchmarks are positioned to the left- and right-hand sides of the dashed line, respectively.}
  \Description{}
  \label{fig:vis}
\end{figure*}

\noindent \textbf{Comparison with SOTA.} We evaluate our method against existing approaches on six established OVSS benchmarks. As shown in Table~\ref{tab:ov}, consistent with state-of-the-art methods (e.g., SAN~\citep{xu2023side}, SED~\citep{xie2023sed}, CAT-Seg~\citep{cho2023cat}, and RPN~\citep{li2024relationship}), our approach achieves competitive performance using only CLIP adaptations, without additional backbone networks or supplementary training datasets. When using the CLIP-B/16 model, our method achieves state-of-the-art performance on four benchmarks: PC-459 (19.2\% mIoU), A-150 (32.1\% mIoU), PC-59 (57.7\% mIoU), and PAS-21 (77.4\% mIoU). On the remaining two benchmarks, it ranks second, with only a marginal 0.1\% mIoU gap to the top performer. When employing the CLIP-L/14 model, our method achieves state-of-the-art performance across all six benchmarks, with improvements of +0.4\%, +0.3\%, +0.4\%, +0.6\%, and +0.2\% mIoU over the second-best approaches on five benchmarks, and parity on the sixth. As illustrated in Figure~\ref{fig:vis}, we showcase qualitative results for the A-150 and PC-459 benchmarks, highlighting significant performance gains of our method against state-of-the-art approaches.

\noindent \textbf{Zero-shot comparison.} We further explore the discriminative generalization of the proposed method for unseen categories within inductive learning. Following the existing zero-shot setting~\cite{li2024relationship,zhou2023zegclip,jiao2023learning}, as shown in the table ~\ref{tab:zs}, our method shows significant performance both on seen and unseen categories (improved by 0.8\% hIoU). Extended experiments across multiple benchmarks are provided in the appendix. We further employ agent token visualization to interpretively validate our framework's capability in latent semantic perception for novel categories in the appendix.

\begin{table}[!t]
\Large
\centering
\caption{Effect of different components (unit: \%). We perform a ablation study by incrementally integrating core components, evaluating their incremental impact on performance.}
\label{tab:ablation_1}
\resizebox{\linewidth}{!}{
    \begin{tabular}{ll!{\vrule height 10pt}cccccc}
    \toprule
    &Components & A-847 & PC-459 & A-150 & PC-59 & PAS-20 & PAS-21\\
    \midrule\midrule
    \textbf{(I)} & Cross Attn. & 5.7 & 11.6 & 20.1 & 47.3 & 90.0 & 64.9\\
    \midrule
    \textbf{(II)} & Agent Attn.  & 10.9 & 17.0 & 28.4 & 53.6 & 94.5 & 72.7\\
    \textbf{(III)} &\textbf{(II)} + Visual Pool.  & 10.9 & 17.5 & 29.7 & 54.4 & 94.6 & 74.1\\
    \textbf{(IV)} &\textbf{(II)} + Textual Pool.  & 10.9 & 17.9 & 29.5 & 53.8 & 94.6 & 72.9\\
    \textbf{(V)} &\textbf{(II)} + Agent Pool. & 11.2 & 18.5 & 30.9 & 55.8 & 94.7 & 75.3\\
    \textbf{(VI)} &\textbf{(V)} + Diff Attn.  & \underline{11.5} & \underline{19.1} & \underline{31.9} & \underline{56.9} & \underline{94.9} & \underline{77.1}\\
    \hlrow
    \textbf{(VII)} &\textbf{(VI)} + $\mathcal{L}_{\text{align}}$  & \textbf{11.9} & \textbf{19.2}& \textbf{32.1}& \textbf{57.7}& \textbf{95.1}&\textbf{77.4}\\
    \bottomrule
    \end{tabular}}
\end{table}
\begin{table}[!t]
\Large
\centering
\caption{Effect of different strategy for the agent selection, the agent pooling and the cost matrix (unit: \%). ``Random.'': The agent tokens are randomly sampled from the \textit{value} matrix $\bm V$. ``Learnable.'': The agent tokens are initialized as trainable parameters and optimized end-to-end. ``Cosine.'': The agent tokens are selected via a cosine similarity-based affinity matrix. ``O.T.'': Optimal Transport. ``MAE.'': Mean Absolute Error. ``MSE.'': Mean Square Error. ``DotMat.'': Matrix Product as Equation~\ref{eq:C}. ``$\gamma$.'': Initializing a single learnable scalar parameter.}
\label{tab:ablation_2}
\resizebox{\linewidth}{!}{
    \begin{tabular}{ll|cccccc}
    \hline
    & Variants & A-847 & PC-459 & A-150 & PC-59 & PAS-20 & PAS-21\\
    \hline
    \multicolumn{8}{c}{\scalebox{1.0}{agent selection}}\\
    \hline
    \textbf{(I)} & Random. & 7.3 & 13.6 & 24.5 & 50.7 & 93.6 & 71.1\\
    \textbf{(II)} & Learnable.  & 8.1 & 14.3 & 26.9 & 53.7 & 94.2 & 72.6\\
    \textbf{(III)} & Cosine.  & \underline{10.1} & \underline{18.7} & \underline{30.3} & \underline{56.5} & \underline{94.8} & \underline{75.4}\\
    \textbf{(IV)} & O.T.  & {9.3} & {17.8} & {30.1} & {55.9} & 94.3 & 75.2\\
    \hlrow
    \textbf{(VI)} &\textbf{(III)} + \textbf{(IV)}  & \textbf{11.9} & \textbf{19.2}& \textbf{32.1}& \textbf{57.7}& \textbf{95.1}&\textbf{77.4}\\
    \hline
    \multicolumn{8}{c}{\scalebox{1.0}{cost matrix}}\\
    \hline
    \textbf{(I)} & MAE. & \underline{10.3} & 18.6 & 31.8 & \bf 57.7 & \underline{94.4} & \underline{76.9}\\
    \textbf{(II)} & MSE. & 10.2 & \underline{18.7} & \underline{31.9} & 57.3 & 94.1 & 76.8\\
    \hlrow
    \textbf{(III)} & DotMat. & \textbf{11.9} & \textbf{19.2}& \textbf{32.1}& \textbf{57.7}& \textbf{95.1}&\textbf{77.4}\\
    \hline
    \multicolumn{8}{c}{\scalebox{1.0}{agent pooling}}\\
    \hline
    \textbf{(I)} & $\gamma$. & \underline{11.8} & \underline{18.9} & \underline{31.9} & \underline{56.6} & \underline{94.8} & \underline{77.0}\\
    \hlrow
    \textbf{(II)} & $\gamma_v+\gamma_t$.  & \textbf{11.9} & \textbf{19.2}& \textbf{32.1}& \textbf{57.7}& \textbf{95.1}&\textbf{77.4}\\
    \hline
    \end{tabular}}
\end{table}
\begin{figure}[t]
  \centering
  \includegraphics[width=\linewidth]{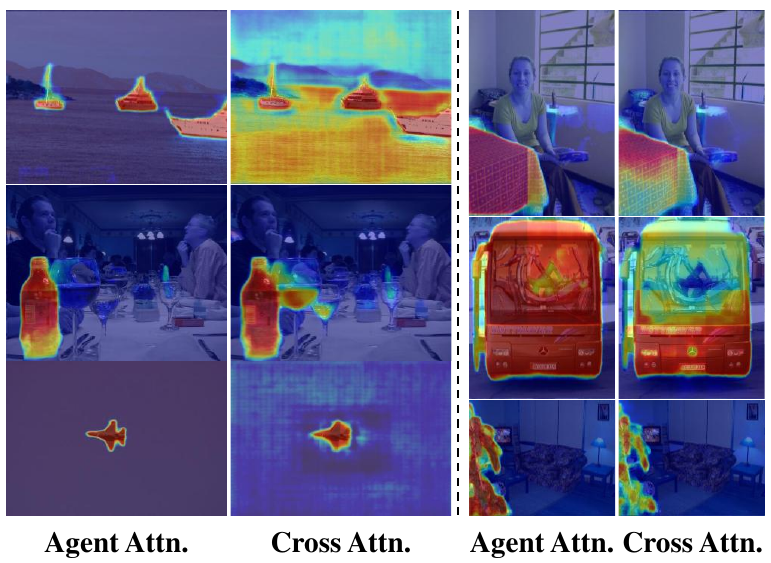}
  \caption{Comparative analysis of attention heatmaps. Our agent attention mechanism enables fine-grained activation within semantically salient regions while suppressing spurious activations, outperforming standard cross-attention in background noise suppression and target-specific saliency enhancement.}
  \Description{}
  \label{fig:heatmap}
\end{figure}

\subsection{Ablation Study}
\noindent \textbf{Component analysis.} To validate the effectiveness of the proposed components, we establish a baseline by directly applying cross-attention (Equation~\ref{eq:cross}) to the cross-layer visual embeddings, then incrementally integrate our components to measure performance gains. As shown in Table~\ref{tab:ablation_1}, substituting cross-attention with agent attention achieves absolute mIoU gains across all benchmarks: A-847 (+5.2\% mIoU), PC-459 (+5.4\% mIoU), A-150 (+8.3\% mIoU), PC-59 (+6.3\% mIoU), PAS-20 (+4.5\% mIoU), and PAS-21 (+7.8\% mIoU). Figure~\ref{fig:heatmap} provides a comparative visualization of attention heatmaps, demonstrating that agent attention produces semantically coherent activation maps with superior noise-filtering capabilities. For instance, for the ``table'', ``plant'', and ``bus'' categories, our activation maps demonstrate a more comprehensive focus on objective regions compared to cross-attention; for ``boat'', ``bottle'', and ``airplane'' categories, our method suppresses the background saliency while sharply highlighting the semantic objectives. These results confirm that our agent attention mediates vision-language interactions through our agent tokens—diverging from direct cross-attention—by selectively amplifying latent semantics to enhance objective saliency and suppress irrelevant distractors. Building upon our agent attention component, we incrementally introduce the agent pooling component, where ``Visual Pool.'' and ``Textual Pool.'' denote the visual and textual branches of the agent pooling component, respectively. Following this integration, we construct multi-modal agent tokens, further enhancing performance. Additionally, leveraging the advantages of the differential attention mechanism in noise suppression and context-aware attention refinement, we replace the traditional $\softmax$ attention within agent attention, thereby achieving additional performance gains.

\begin{figure}[t]
  \centering
  \includegraphics[width=\linewidth]{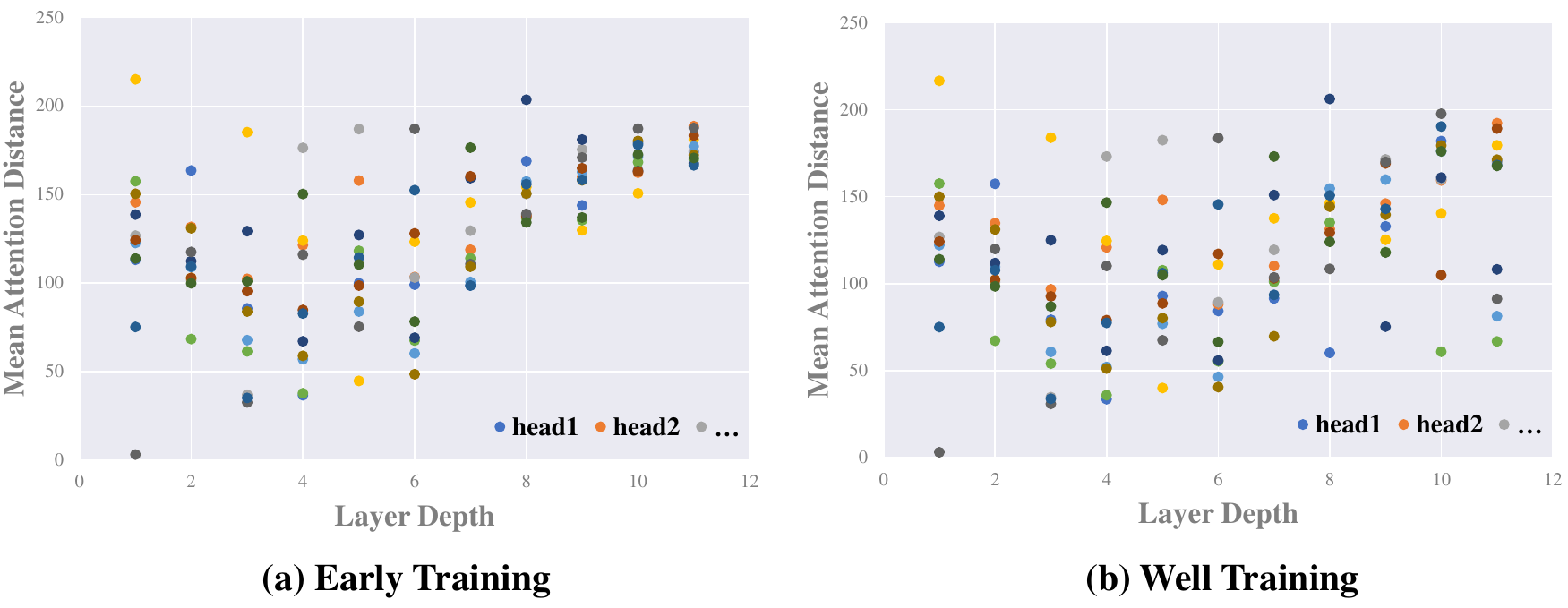}
  \caption{Mean Attention Distance (MAD) analysis of each attention head in visual layers across training phases. Low MAD indicates that the attention is concentrated on adjacent regions, which is suitable for capturing local features such as textures and edges. High MAD reflects that the attention is dispersed over distant positions, enabling the capture of global semantics like object structures and contextual relationships.}
  \Description{}
  \label{fig:mad}
\end{figure}

\noindent \textbf{Agent selection analysis.} Selecting appropriate agent tokens is the foundational and critical first step in our methodology. We systematically explore four agent selection strategies, broadly categorized into two classes: (1) Prior-free: Encompassing random selection and learnable approaches; (2) Prior-guided: Including cosine similarity-based and optimal transport-based methods. As shown in Table~\ref{tab:ablation_2}, prior-free agent selection strategies exhibit inferior performance compared to prior-guided approaches. This indicates that latent semantic tokens are more cryptically distributed in high-dimensional feature spaces, making them difficult to capture through gradient-based learning mechanisms without prior guidance. In contrast, prior-guided strategies inject domain-specific inductive biases, constraining the hypothesis space during optimization and facilitating precise localization of latent semantics.

\noindent \textbf{Cost matrix analysis.} We investigate the effect of different cost matrices in optimal transport on performance, including Euclidean distance-based approaches (e.g., mean absolute difference and mean squared difference) and dot product-based methods (i.e., Equation~\ref{eq:C}). As shown in Table~\ref{tab:ablation_2}, the dot product-based approach achieves optimal performance, which aligns with the intuitive understanding of cost definitions in multi-modal alignment.

\noindent \textbf{Agent pooling analysis.} We analyze the effect of varying scalar configurations in Equation~\ref{eq:agent} on performance. As shown in Table~\ref{tab:ablation_2}, the results demonstrate that a mutual learning strategy (i.e., Equation~\ref{eq:gamma}) yields superior performance compared to using a single learnable parameter.

\noindent \textbf{Attention matrices analysis.} In this part, we investigate two critical aspects: (1) selecting which attention matrix serves as the input to optimal transport for constructing the semantic affinity matrix, and (2) determining which attention matrix should be prioritized during agent selection. As shown in Table~\ref{tab:ablation_3}, our experiments reveal that aligning the \textit{key} matrix with text tokens yields a more semantically relevant affinity matrix, while leveraging the \textit{value} matrix as the target for agent selection better encapsulates latent semantics. Additionally, we observe a significant performance degradation when employing the \textit{query} matrix as the target for agent selection. This is likely attributed to the \textit{query} matrix's inherent limitations in serving as an effective intermediary for multi-modal interactions.

\noindent \textbf{Semantic refinement analysis.} Our method also demonstrates exceptional performance in semantic refinement. As illustrated in Figure~\ref{fig:mad}, we compare the Mean Attention Distance (MAD) of multi-head attention in the visual layers during the early training phase and the well-training phase. This metric reflects the receptive field of the visual layers, i.e., the degree of information aggregation. For OVSS task, focusing on high-frequency features such as edges and textures is critical, necessitating lower MAD values in visual layer feature representations. Furthermore, deeper features inherently exhibit higher levels of semantic abstraction; thus, as shown in Figure~\ref{fig:mad} (a), deeper layers in CLIP-like image-level encoder typically display larger MAD values (i.e., global attention). At the well-training stage (Figure~\ref{fig:mad} (b)), our deep-layer features demonstrate a broader range of MAD values, indicating that our method simultaneously attends to holistic semantics and fine-grained details.

\begin{table}[!t]
\Large
\centering
\caption{Effect of different interactions with the attention matrix during the affinity matrix construction and the agent selection (unit: \%). ``From $\bm Q$ to $\bm K$'' denotes calculating the affinity between $\bm Q$ and the textual tokens and selecting the agent tokens within $\bm K$.}
\label{tab:ablation_3}
\resizebox{\linewidth}{!}{
    \begin{tabular}{lc!{\vrule height 10pt}cccccc}
    \toprule
    & Variants & A-847 & PC-459 & A-150 & PC-59 & PAS-20 & PAS-21\\
    \midrule\midrule
    & \textit{From} \{\,\} \textit{to} ${\bm V}$  &  &  &  &  &  & \\ 
    \midrule
    \textbf{(I)} & $\bm Q$ & 10.4 & 18.3 & 31.4 & 57.1 & 94.0 & 76.7\\
    \textbf{(II)} & $\bm V$ & \underline{11.2} & \underline{18.6} & \underline{31.6} & \underline{57.3} & \underline{94.6} & \underline{77.1}\\
    \hlrow
    \textbf{(III)} & $\bm K$  & \textbf{11.9} & \textbf{19.2}& \textbf{32.1}& \textbf{57.7}& \textbf{95.1}&\textbf{77.4}\\
    \midrule
    & \textit{From} ${\bm K}$ \textit{to} \{\,\} &  &  &  &  &  & \\
    \midrule
    \textbf{(IV)} & $\bm Q$ & 5.8 & 12.6 & 21.3 & 49.4 & 91.2 & 65.7\\
    \textbf{(II)} & $\bm K$ & \underline{11.3} & \underline{18.6} & \underline{32.1} & \underline{57.3} & \underline{95.0} & \underline{77.1}\\
    \hlrow
    \textbf{(III)} & $\bm V$  & \textbf{11.9} & \textbf{19.2}& \textbf{32.1}& \textbf{57.7}& \textbf{95.1}&\textbf{77.4}\\
    \bottomrule
    \end{tabular}}
\end{table}

\section{Conclusion}
In this study, we address a critical yet underexplored issue in VLM-based OVSS: preserving latent semantic generalization in VLMs under inductive supervision while preventing semantic dissipation. We first conduct a probing experiment that reveals the distribution patterns and dynamic propagation characteristics of latent semantics in VLM's high-dimensional embedding space. Building on these findings, we systematically construct the X-Agent framework—an enhanced OVSS architecture for persistent latent semantic awareness—comprising three innovative components: (1) agent selection component: constructs latent semantic-aware ``agent'' by excavating latent unseen semantic tokens in VLMs; (2) agent pooling component: establishes cross-modal global contextual awareness to strengthen ``agent'' semantic representational capacity; (3) agent attention component: integrates semantic ``agent'' into multi-modal cross-attention mechanisms for adaptive enhancement of latent semantic saliency. Comprehensive experimental evaluations demonstrate breakthrough improvements in both latent semantic perception granularity and interference suppression capability. Furthermore, our approach can be conceptualized as a parameter-efficient fine-tuning (PEFT) paradigm that directly modulates the core attention mechanisms of VLM to steer the semantic optimization process towards desired evolutionary trajectories, providing novel optimization insights for large-scale model fine-tuning.

\section*{Acknowledgments}
This work was supported in part by National Natural Science Foundation of China under Grant 62376233, 62176224, 62176092, 62222602, 62306165, and 62306165, in part by Science and Technology on Sonar Laboratory under grant 2024-JCJQ-LB-32/07, in part by China Academy of RailwaySciences under Grant 2023Y1357, in part by Xiaomi Young Talents Program award, in part by Natural Science Foundation of Shanghai under Grant 23ZR1420400; and in part by Natural Science Foundation of Chongqing under Grant CSTB2023NSCQ-JQX0007.

\bibliographystyle{ACM-Reference-Format}
\bibliography{main}


\begin{thebibliography}{61}


\ifx \showCODEN    \undefined \def \showCODEN     #1{\unskip}     \fi
\ifx \showISBNx    \undefined \def \showISBNx     #1{\unskip}     \fi
\ifx \showISBNxiii \undefined \def \showISBNxiii  #1{\unskip}     \fi
\ifx \showISSN     \undefined \def \showISSN      #1{\unskip}     \fi
\ifx \showLCCN     \undefined \def \showLCCN      #1{\unskip}     \fi
\ifx \shownote     \undefined \def \shownote      #1{#1}          \fi
\ifx \showarticletitle \undefined \def \showarticletitle #1{#1}   \fi
\ifx \showURL      \undefined \def \showURL       {\relax}        \fi
\providecommand\bibfield[2]{#2}
\providecommand\bibinfo[2]{#2}
\providecommand\natexlab[1]{#1}
\providecommand\showeprint[2][]{arXiv:#2}

\bibitem[Bucher et~al\mbox{.}(2019)]%
        {bucher2019zero}
\bibfield{author}{\bibinfo{person}{Maxime Bucher}, \bibinfo{person}{Tuan-Hung Vu}, \bibinfo{person}{Matthieu Cord}, {and} \bibinfo{person}{Patrick P{\'e}rez}.} \bibinfo{year}{2019}\natexlab{}.
\newblock \showarticletitle{Zero-shot semantic segmentation}.
\newblock \bibinfo{journal}{\emph{Advances in Neural Information Processing Systems}}  \bibinfo{volume}{32} (\bibinfo{year}{2019}).
\newblock


\bibitem[Caesar et~al\mbox{.}(2018)]%
        {caesar2018cvpr}
\bibfield{author}{\bibinfo{person}{Holger Caesar}, \bibinfo{person}{Jasper Uijlings}, {and} \bibinfo{person}{Vittorio Ferrari}.} \bibinfo{year}{2018}\natexlab{}.
\newblock \showarticletitle{COCO-Stuff: Thing and stuff classes in context}. In \bibinfo{booktitle}{\emph{Computer Vision and Pattern Recognition (CVPR), 2018 IEEE conference on}}. IEEE.
\newblock


\bibitem[Chen et~al\mbox{.}(2020)]%
        {chen2020learning}
\bibfield{author}{\bibinfo{person}{Guangyao Chen}, \bibinfo{person}{Limeng Qiao}, \bibinfo{person}{Yemin Shi}, \bibinfo{person}{Peixi Peng}, \bibinfo{person}{Jia Li}, \bibinfo{person}{Tiejun Huang}, \bibinfo{person}{Shiliang Pu}, {and} \bibinfo{person}{Yonghong Tian}.} \bibinfo{year}{2020}\natexlab{}.
\newblock \showarticletitle{Learning open set network with discriminative reciprocal points}. In \bibinfo{booktitle}{\emph{Computer Vision--ECCV 2020: 16th European Conference, Glasgow, UK, August 23--28, 2020, Proceedings, Part III 16}}. Springer, \bibinfo{pages}{507--522}.
\newblock


\bibitem[Cheng et~al\mbox{.}(2021)]%
        {cheng2021sign}
\bibfield{author}{\bibinfo{person}{Jiaxin Cheng}, \bibinfo{person}{Soumyaroop Nandi}, \bibinfo{person}{Prem Natarajan}, {and} \bibinfo{person}{Wael Abd-Almageed}.} \bibinfo{year}{2021}\natexlab{}.
\newblock \showarticletitle{Sign: Spatial-information incorporated generative network for generalized zero-shot semantic segmentation}. In \bibinfo{booktitle}{\emph{Proceedings of the IEEE/CVF International Conference on Computer Vision}}. \bibinfo{pages}{9556--9566}.
\newblock


\bibitem[Cho et~al\mbox{.}(2023)]%
        {cho2023cat}
\bibfield{author}{\bibinfo{person}{Seokju Cho}, \bibinfo{person}{Heeseong Shin}, \bibinfo{person}{Sunghwan Hong}, \bibinfo{person}{Seungjun An}, \bibinfo{person}{Seungjun Lee}, \bibinfo{person}{Anurag Arnab}, \bibinfo{person}{Paul~Hongsuck Seo}, {and} \bibinfo{person}{Seungryong Kim}.} \bibinfo{year}{2023}\natexlab{}.
\newblock \showarticletitle{Cat-seg: Cost aggregation for open-vocabulary semantic segmentation}.
\newblock \bibinfo{journal}{\emph{arXiv preprint arXiv:2303.11797}} (\bibinfo{year}{2023}).
\newblock


\bibitem[Cuturi(2013)]%
        {cuturi2013sinkhorn}
\bibfield{author}{\bibinfo{person}{Marco Cuturi}.} \bibinfo{year}{2013}\natexlab{}.
\newblock \showarticletitle{Sinkhorn distances: Lightspeed computation of optimal transport}.
\newblock \bibinfo{journal}{\emph{Advances in Neural Information Processing Systems}}  \bibinfo{volume}{26} (\bibinfo{year}{2013}).
\newblock


\bibitem[Ding et~al\mbox{.}(2022)]%
        {ding2022decoupling}
\bibfield{author}{\bibinfo{person}{Jian Ding}, \bibinfo{person}{Nan Xue}, \bibinfo{person}{Gui-Song Xia}, {and} \bibinfo{person}{Dengxin Dai}.} \bibinfo{year}{2022}\natexlab{}.
\newblock \showarticletitle{Decoupling zero-shot semantic segmentation}. In \bibinfo{booktitle}{\emph{Proceedings of the IEEE/CVF Conference on Computer Vision and Pattern Recognition}}. \bibinfo{pages}{11583--11592}.
\newblock


\bibitem[Everingham et~al\mbox{.}({[n.\,d.]})]%
        {pascal-voc-2012}
\bibfield{author}{\bibinfo{person}{M. Everingham}, \bibinfo{person}{L. Van~Gool}, \bibinfo{person}{C.~K.~I. Williams}, \bibinfo{person}{J. Winn}, {and} \bibinfo{person}{A. Zisserman}.} \bibinfo{year}{[n.\,d.]}\natexlab{}.
\newblock \bibinfo{title}{The {PASCAL} {V}isual {O}bject {C}lasses {C}hallenge 2012 {(VOC2012)} {R}esults}.
\newblock \bibinfo{howpublished}{http://www.pascal-network.org/challenges/VOC/voc2012/workshop/index.html}.
\newblock


\bibitem[Fini et~al\mbox{.}(2021)]%
        {uno}
\bibfield{author}{\bibinfo{person}{Enrico Fini}, \bibinfo{person}{Enver Sangineto}, \bibinfo{person}{St{\'e}phane Lathuili{\`e}re}, \bibinfo{person}{Zhun Zhong}, \bibinfo{person}{Moin Nabi}, {and} \bibinfo{person}{Elisa Ricci}.} \bibinfo{year}{2021}\natexlab{}.
\newblock \showarticletitle{A unified objective for novel class discovery}. In \bibinfo{booktitle}{\emph{Proceedings of the IEEE/CVF International Conference on Computer Vision}}. \bibinfo{pages}{9284--9292}.
\newblock


\bibitem[Ghiasi et~al\mbox{.}(2022)]%
        {ghiasi2022scaling}
\bibfield{author}{\bibinfo{person}{Golnaz Ghiasi}, \bibinfo{person}{Xiuye Gu}, \bibinfo{person}{Yin Cui}, {and} \bibinfo{person}{Tsung-Yi Lin}.} \bibinfo{year}{2022}\natexlab{}.
\newblock \showarticletitle{Scaling open-vocabulary image segmentation with image-level labels}. In \bibinfo{booktitle}{\emph{European Conference on Computer Vision}}. Springer, \bibinfo{pages}{540--557}.
\newblock


\bibitem[Gu et~al\mbox{.}(2020)]%
        {gu2020context}
\bibfield{author}{\bibinfo{person}{Zhangxuan Gu}, \bibinfo{person}{Siyuan Zhou}, \bibinfo{person}{Li Niu}, \bibinfo{person}{Zihan Zhao}, {and} \bibinfo{person}{Liqing Zhang}.} \bibinfo{year}{2020}\natexlab{}.
\newblock \showarticletitle{Context-aware feature generation for zero-shot semantic segmentation}. In \bibinfo{booktitle}{\emph{Proceedings of the 28th ACM International Conference on Multimedia}}. \bibinfo{pages}{1921--1929}.
\newblock


\bibitem[Han et~al\mbox{.}(2024)]%
        {han2024agent}
\bibfield{author}{\bibinfo{person}{Dongchen Han}, \bibinfo{person}{Tianzhu Ye}, \bibinfo{person}{Yizeng Han}, \bibinfo{person}{Zhuofan Xia}, \bibinfo{person}{Siyuan Pan}, \bibinfo{person}{Pengfei Wan}, \bibinfo{person}{Shiji Song}, {and} \bibinfo{person}{Gao Huang}.} \bibinfo{year}{2024}\natexlab{}.
\newblock \showarticletitle{Agent attention: On the integration of softmax and linear attention}. In \bibinfo{booktitle}{\emph{European Conference on Computer Vision}}. Springer, \bibinfo{pages}{124--140}.
\newblock


\bibitem[Han et~al\mbox{.}(2020)]%
        {han20automatically}
\bibfield{author}{\bibinfo{person}{Kai Han}, \bibinfo{person}{Sylvestre-Alvise Rebuffi}, \bibinfo{person}{Sebastien Ehrhardt}, \bibinfo{person}{Andrea Vedaldi}, {and} \bibinfo{person}{Andrew Zisserman}.} \bibinfo{year}{2020}\natexlab{}.
\newblock \showarticletitle{Automatically Discovering and Learning New Visual Categories with Ranking Statistics}.
\newblock


\bibitem[Han et~al\mbox{.}(2021)]%
        {han21autonovel}
\bibfield{author}{\bibinfo{person}{Kai Han}, \bibinfo{person}{Sylvestre-Alvise Rebuffi}, \bibinfo{person}{Sebastien Ehrhardt}, \bibinfo{person}{Andrea Vedaldi}, {and} \bibinfo{person}{Andrew Zisserman}.} \bibinfo{year}{2021}\natexlab{}.
\newblock \showarticletitle{AutoNovel: Automatically Discovering and Learning Novel Visual Categories}.
\newblock  (\bibinfo{year}{2021}).
\newblock


\bibitem[Han et~al\mbox{.}(2019)]%
        {han2019learning}
\bibfield{author}{\bibinfo{person}{Kai Han}, \bibinfo{person}{Andrea Vedaldi}, {and} \bibinfo{person}{Andrew Zisserman}.} \bibinfo{year}{2019}\natexlab{}.
\newblock \showarticletitle{Learning to Discover Novel Visual Categories via Deep Transfer Clustering}.
\newblock


\bibitem[He et~al\mbox{.}(2023)]%
        {he2023primitive}
\bibfield{author}{\bibinfo{person}{Shuting He}, \bibinfo{person}{Henghui Ding}, {and} \bibinfo{person}{Wei Jiang}.} \bibinfo{year}{2023}\natexlab{}.
\newblock \showarticletitle{Primitive generation and semantic-related alignment for universal zero-shot segmentation}. In \bibinfo{booktitle}{\emph{Proceedings of the IEEE/CVF Conference on Computer Vision and Pattern Recognition}}. \bibinfo{pages}{11238--11247}.
\newblock


\bibitem[Hsu et~al\mbox{.}(2018)]%
        {Hsu18_L2C}
\bibfield{author}{\bibinfo{person}{Yen-Chang Hsu}, \bibinfo{person}{Zhaoyang Lv}, {and} \bibinfo{person}{Zsolt Kira}.} \bibinfo{year}{2018}\natexlab{}.
\newblock \showarticletitle{Learning to cluster in order to transfer across domains and tasks}.
\newblock


\bibitem[Hsu et~al\mbox{.}(2019)]%
        {Hsu19_MCL}
\bibfield{author}{\bibinfo{person}{Yen-Chang Hsu}, \bibinfo{person}{Zhaoyang Lv}, \bibinfo{person}{Joel Schlosser}, \bibinfo{person}{Phillip Odom}, {and} \bibinfo{person}{Zsolt Kira}.} \bibinfo{year}{2019}\natexlab{}.
\newblock \showarticletitle{Multi-class classification without multi-class labels}.
\newblock


\bibitem[Jia et~al\mbox{.}(2021b)]%
        {jia2021scaling}
\bibfield{author}{\bibinfo{person}{Chao Jia}, \bibinfo{person}{Yinfei Yang}, \bibinfo{person}{Ye Xia}, \bibinfo{person}{Yi-Ting Chen}, \bibinfo{person}{Zarana Parekh}, \bibinfo{person}{Hieu Pham}, \bibinfo{person}{Quoc Le}, \bibinfo{person}{Yun-Hsuan Sung}, \bibinfo{person}{Zhen Li}, {and} \bibinfo{person}{Tom Duerig}.} \bibinfo{year}{2021}\natexlab{b}.
\newblock \showarticletitle{Scaling up visual and vision-language representation learning with noisy text supervision}. In \bibinfo{booktitle}{\emph{International conference on machine learning}}. PMLR, \bibinfo{pages}{4904--4916}.
\newblock


\bibitem[Jia et~al\mbox{.}(2021a)]%
        {wta}
\bibfield{author}{\bibinfo{person}{Xuhui Jia}, \bibinfo{person}{Kai Han}, \bibinfo{person}{Yukun Zhu}, {and} \bibinfo{person}{Bradley Green}.} \bibinfo{year}{2021}\natexlab{a}.
\newblock \showarticletitle{Joint representation learning and novel category discovery on single-and multi-modal data}. In \bibinfo{booktitle}{\emph{Proceedings of the IEEE/CVF International Conference on Computer Vision}}. \bibinfo{pages}{610--619}.
\newblock


\bibitem[Jiao et~al\mbox{.}(2023)]%
        {jiao2023learning}
\bibfield{author}{\bibinfo{person}{Siyu Jiao}, \bibinfo{person}{Yunchao Wei}, \bibinfo{person}{Yaowei Wang}, \bibinfo{person}{Yao Zhao}, {and} \bibinfo{person}{Humphrey Shi}.} \bibinfo{year}{2023}\natexlab{}.
\newblock \showarticletitle{Learning mask-aware clip representations for zero-shot segmentation}.
\newblock \bibinfo{journal}{\emph{Advances in Neural Information Processing Systems}}  \bibinfo{volume}{36} (\bibinfo{year}{2023}), \bibinfo{pages}{35631--35653}.
\newblock


\bibitem[Kwon et~al\mbox{.}(2023)]%
        {kwon2023probabilistic}
\bibfield{author}{\bibinfo{person}{Hyeongjun Kwon}, \bibinfo{person}{Taeyong Song}, \bibinfo{person}{Somi Jeong}, \bibinfo{person}{Jin Kim}, \bibinfo{person}{Jinhyun Jang}, {and} \bibinfo{person}{Kwanghoon Sohn}.} \bibinfo{year}{2023}\natexlab{}.
\newblock \showarticletitle{Probabilistic Prompt Learning for Dense Prediction}. In \bibinfo{booktitle}{\emph{Proceedings of the IEEE/CVF Conference on Computer Vision and Pattern Recognition}}. \bibinfo{pages}{6768--6777}.
\newblock


\bibitem[Li et~al\mbox{.}(2022)]%
        {li2022language}
\bibfield{author}{\bibinfo{person}{Boyi Li}, \bibinfo{person}{Kilian~Q Weinberger}, \bibinfo{person}{Serge Belongie}, \bibinfo{person}{Vladlen Koltun}, {and} \bibinfo{person}{Ren{\'e} Ranftl}.} \bibinfo{year}{2022}\natexlab{}.
\newblock \showarticletitle{Language-driven semantic segmentation}.
\newblock \bibinfo{journal}{\emph{arXiv preprint arXiv:2201.03546}} (\bibinfo{year}{2022}).
\newblock


\bibitem[Li et~al\mbox{.}(2024)]%
        {li2024relationship}
\bibfield{author}{\bibinfo{person}{Jiahao Li}, \bibinfo{person}{Yang Lu}, \bibinfo{person}{Yuan Xie}, {and} \bibinfo{person}{Yanyun Qu}.} \bibinfo{year}{2024}\natexlab{}.
\newblock \showarticletitle{Relationship Prompt Learning is Enough for Open-Vocabulary Semantic Segmentation}.
\newblock \bibinfo{journal}{\emph{Advances in Neural Information Processing Systems}}  \bibinfo{volume}{37} (\bibinfo{year}{2024}), \bibinfo{pages}{74298--74324}.
\newblock


\bibitem[Li et~al\mbox{.}(2020)]%
        {li2020consistent}
\bibfield{author}{\bibinfo{person}{Peike Li}, \bibinfo{person}{Yunchao Wei}, {and} \bibinfo{person}{Yi Yang}.} \bibinfo{year}{2020}\natexlab{}.
\newblock \showarticletitle{Consistent structural relation learning for zero-shot segmentation}.
\newblock \bibinfo{journal}{\emph{Advances in Neural Information Processing Systems}}  \bibinfo{volume}{33} (\bibinfo{year}{2020}), \bibinfo{pages}{10317--10327}.
\newblock


\bibitem[Liang et~al\mbox{.}(2022)]%
        {liang2022open}
\bibfield{author}{\bibinfo{person}{Feng Liang}, \bibinfo{person}{Bichen Wu}, \bibinfo{person}{Xiaoliang Dai}, \bibinfo{person}{Kunpeng Li}, \bibinfo{person}{Yinan Zhao}, \bibinfo{person}{Hang Zhang}, \bibinfo{person}{Peizhao Zhang}, \bibinfo{person}{Peter Vajda}, {and} \bibinfo{person}{Diana Marculescu}.} \bibinfo{year}{2022}\natexlab{}.
\newblock \showarticletitle{Open-vocabulary semantic segmentation with mask-adapted clip}.
\newblock \bibinfo{journal}{\emph{arXiv preprint arXiv:2210.04150}} (\bibinfo{year}{2022}).
\newblock


\bibitem[Liang et~al\mbox{.}(2023)]%
        {liang2023open}
\bibfield{author}{\bibinfo{person}{Feng Liang}, \bibinfo{person}{Bichen Wu}, \bibinfo{person}{Xiaoliang Dai}, \bibinfo{person}{Kunpeng Li}, \bibinfo{person}{Yinan Zhao}, \bibinfo{person}{Hang Zhang}, \bibinfo{person}{Peizhao Zhang}, \bibinfo{person}{Peter Vajda}, {and} \bibinfo{person}{Diana Marculescu}.} \bibinfo{year}{2023}\natexlab{}.
\newblock \showarticletitle{Open-vocabulary semantic segmentation with mask-adapted clip}. In \bibinfo{booktitle}{\emph{Proceedings of the IEEE/CVF Conference on Computer Vision and Pattern Recognition}}. \bibinfo{pages}{7061--7070}.
\newblock


\bibitem[Liu et~al\mbox{.}(2020)]%
        {liu2020few}
\bibfield{author}{\bibinfo{person}{Bo Liu}, \bibinfo{person}{Hao Kang}, \bibinfo{person}{Haoxiang Li}, \bibinfo{person}{Gang Hua}, {and} \bibinfo{person}{Nuno Vasconcelos}.} \bibinfo{year}{2020}\natexlab{}.
\newblock \showarticletitle{Few-shot open-set recognition using meta-learning}. In \bibinfo{booktitle}{\emph{Proceedings of the IEEE/CVF Conference on Computer Vision and Pattern Recognition}}. \bibinfo{pages}{8798--8807}.
\newblock


\bibitem[Liu et~al\mbox{.}(2017)]%
        {liu2017incremental}
\bibfield{author}{\bibinfo{person}{Juncheng Liu}, \bibinfo{person}{Zhouhui Lian}, \bibinfo{person}{Yi Wang}, {and} \bibinfo{person}{Jianguo Xiao}.} \bibinfo{year}{2017}\natexlab{}.
\newblock \showarticletitle{Incremental kernel null space discriminant analysis for novelty detection}. In \bibinfo{booktitle}{\emph{Proceedings of the IEEE Conference on Computer Vision and Pattern Recognition}}. \bibinfo{pages}{792--800}.
\newblock


\bibitem[Liu et~al\mbox{.}(2023)]%
        {liu2023delving}
\bibfield{author}{\bibinfo{person}{Xinyu Liu}, \bibinfo{person}{Beiwen Tian}, \bibinfo{person}{Zhen Wang}, \bibinfo{person}{Rui Wang}, \bibinfo{person}{Kehua Sheng}, \bibinfo{person}{Bo Zhang}, \bibinfo{person}{Hao Zhao}, {and} \bibinfo{person}{Guyue Zhou}.} \bibinfo{year}{2023}\natexlab{}.
\newblock \showarticletitle{Delving into Shape-aware Zero-shot Semantic Segmentation}. In \bibinfo{booktitle}{\emph{Proceedings of the IEEE/CVF Conference on Computer Vision and Pattern Recognition}}. \bibinfo{pages}{2999--3009}.
\newblock


\bibitem[Masana et~al\mbox{.}(2018)]%
        {masana2018metric}
\bibfield{author}{\bibinfo{person}{Marc Masana}, \bibinfo{person}{Idoia Ruiz}, \bibinfo{person}{Joan Serrat}, \bibinfo{person}{Joost van~de Weijer}, {and} \bibinfo{person}{Antonio~M Lopez}.} \bibinfo{year}{2018}\natexlab{}.
\newblock \showarticletitle{Metric learning for novelty and anomaly detection}.
\newblock \bibinfo{journal}{\emph{arXiv preprint arXiv:1808.05492}} (\bibinfo{year}{2018}).
\newblock


\bibitem[Mottaghi et~al\mbox{.}(2014)]%
        {mottaghi_cvpr14}
\bibfield{author}{\bibinfo{person}{Roozbeh Mottaghi}, \bibinfo{person}{Xianjie Chen}, \bibinfo{person}{Xiaobai Liu}, \bibinfo{person}{Nam-Gyu Cho}, \bibinfo{person}{Seong-Whan Lee}, \bibinfo{person}{Sanja Fidler}, \bibinfo{person}{Raquel Urtasun}, {and} \bibinfo{person}{Alan Yuille}.} \bibinfo{year}{2014}\natexlab{}.
\newblock \showarticletitle{The Role of Context for Object Detection and Semantic Segmentation in the Wild}. In \bibinfo{booktitle}{\emph{IEEE Conference on Computer Vision and Pattern Recognition (CVPR)}}.
\newblock


\bibitem[Pastore et~al\mbox{.}(2021)]%
        {pastore2021closer}
\bibfield{author}{\bibinfo{person}{Giuseppe Pastore}, \bibinfo{person}{Fabio Cermelli}, \bibinfo{person}{Yongqin Xian}, \bibinfo{person}{Massimiliano Mancini}, \bibinfo{person}{Zeynep Akata}, {and} \bibinfo{person}{Barbara Caputo}.} \bibinfo{year}{2021}\natexlab{}.
\newblock \showarticletitle{A closer look at self-training for zero-label semantic segmentation}. In \bibinfo{booktitle}{\emph{Proceedings of the IEEE/CVF Conference on Computer Vision and Pattern Recognition}}. \bibinfo{pages}{2693--2702}.
\newblock


\bibitem[Paszke et~al\mbox{.}(2019)]%
        {paszke2019pytorch}
\bibfield{author}{\bibinfo{person}{Adam Paszke}, \bibinfo{person}{Sam Gross}, \bibinfo{person}{Francisco Massa}, \bibinfo{person}{Adam Lerer}, \bibinfo{person}{James Bradbury}, \bibinfo{person}{Gregory Chanan}, \bibinfo{person}{Trevor Killeen}, \bibinfo{person}{Zeming Lin}, \bibinfo{person}{Natalia Gimelshein}, \bibinfo{person}{Luca Antiga}, {et~al\mbox{.}}} \bibinfo{year}{2019}\natexlab{}.
\newblock \showarticletitle{Pytorch: An imperative style, high-performance deep learning library}.
\newblock \bibinfo{journal}{\emph{Advances in neural information processing systems}}  \bibinfo{volume}{32} (\bibinfo{year}{2019}).
\newblock


\bibitem[Qin et~al\mbox{.}(2023)]%
        {qin2023freeseg}
\bibfield{author}{\bibinfo{person}{Jie Qin}, \bibinfo{person}{Jie Wu}, \bibinfo{person}{Pengxiang Yan}, \bibinfo{person}{Ming Li}, \bibinfo{person}{Ren Yuxi}, \bibinfo{person}{Xuefeng Xiao}, \bibinfo{person}{Yitong Wang}, \bibinfo{person}{Rui Wang}, \bibinfo{person}{Shilei Wen}, \bibinfo{person}{Xin Pan}, {et~al\mbox{.}}} \bibinfo{year}{2023}\natexlab{}.
\newblock \showarticletitle{FreeSeg: Unified, Universal and Open-Vocabulary Image Segmentation}. In \bibinfo{booktitle}{\emph{Proceedings of the IEEE/CVF Conference on Computer Vision and Pattern Recognition}}. \bibinfo{pages}{19446--19455}.
\newblock


\bibitem[Radford et~al\mbox{.}(2021)]%
        {radford2021learning}
\bibfield{author}{\bibinfo{person}{Alec Radford}, \bibinfo{person}{Jong~Wook Kim}, \bibinfo{person}{Chris Hallacy}, \bibinfo{person}{Aditya Ramesh}, \bibinfo{person}{Gabriel Goh}, \bibinfo{person}{Sandhini Agarwal}, \bibinfo{person}{Girish Sastry}, \bibinfo{person}{Amanda Askell}, \bibinfo{person}{Pamela Mishkin}, \bibinfo{person}{Jack Clark}, {et~al\mbox{.}}} \bibinfo{year}{2021}\natexlab{}.
\newblock \showarticletitle{Learning transferable visual models from natural language supervision}. In \bibinfo{booktitle}{\emph{International conference on Machine Learning}}. PMLR, \bibinfo{pages}{8748--8763}.
\newblock


\bibitem[Scheirer et~al\mbox{.}(2013)]%
        {Scheirer_2013_TPAMI}
\bibfield{author}{\bibinfo{person}{Walter~J. Scheirer}, \bibinfo{person}{Anderson Rocha}, \bibinfo{person}{Archana Sapkota}, {and} \bibinfo{person}{Terrance~E. Boult}.} \bibinfo{year}{2013}\natexlab{}.
\newblock \showarticletitle{Towards Open Set Recognition}.
\newblock  (\bibinfo{year}{2013}).
\newblock


\bibitem[Shao et~al\mbox{.}(2020)]%
        {shao2020open}
\bibfield{author}{\bibinfo{person}{Rui Shao}, \bibinfo{person}{Pramuditha Perera}, \bibinfo{person}{Pong~C Yuen}, {and} \bibinfo{person}{Vishal~M Patel}.} \bibinfo{year}{2020}\natexlab{}.
\newblock \showarticletitle{Open-set adversarial defense}. In \bibinfo{booktitle}{\emph{European Conference on Computer Vision}}. Springer, \bibinfo{pages}{682--698}.
\newblock


\bibitem[Shen et~al\mbox{.}(2021)]%
        {shen2021conterfactual}
\bibfield{author}{\bibinfo{person}{Feihong Shen}, \bibinfo{person}{Jun Liu}, {and} \bibinfo{person}{Ping Hu}.} \bibinfo{year}{2021}\natexlab{}.
\newblock \showarticletitle{Conterfactual generative zero-shot semantic segmentation}.
\newblock \bibinfo{journal}{\emph{arXiv preprint arXiv:2106.06360}} (\bibinfo{year}{2021}).
\newblock


\bibitem[Shi et~al\mbox{.}(2025)]%
        {shi2025llmformer}
\bibfield{author}{\bibinfo{person}{Hengcan Shi}, \bibinfo{person}{Son~Duy Dao}, {and} \bibinfo{person}{Jianfei Cai}.} \bibinfo{year}{2025}\natexlab{}.
\newblock \showarticletitle{LLMFormer: Large language model for open-vocabulary semantic segmentation}.
\newblock \bibinfo{journal}{\emph{International Journal of Computer Vision}} \bibinfo{volume}{133}, \bibinfo{number}{2} (\bibinfo{year}{2025}), \bibinfo{pages}{742--759}.
\newblock


\bibitem[Shu et~al\mbox{.}(2020)]%
        {shu2020p}
\bibfield{author}{\bibinfo{person}{Yu Shu}, \bibinfo{person}{Yemin Shi}, \bibinfo{person}{Yaowei Wang}, \bibinfo{person}{Tiejun Huang}, {and} \bibinfo{person}{Yonghong Tian}.} \bibinfo{year}{2020}\natexlab{}.
\newblock \showarticletitle{P-odn: Prototype-based open deep network for open set recognition}.
\newblock \bibinfo{journal}{\emph{Scientific reports}} \bibinfo{volume}{10}, \bibinfo{number}{1} (\bibinfo{year}{2020}), \bibinfo{pages}{7146}.
\newblock


\bibitem[Singh et~al\mbox{.}(2022)]%
        {singh2022flava}
\bibfield{author}{\bibinfo{person}{Amanpreet Singh}, \bibinfo{person}{Ronghang Hu}, \bibinfo{person}{Vedanuj Goswami}, \bibinfo{person}{Guillaume Couairon}, \bibinfo{person}{Wojciech Galuba}, \bibinfo{person}{Marcus Rohrbach}, {and} \bibinfo{person}{Douwe Kiela}.} \bibinfo{year}{2022}\natexlab{}.
\newblock \showarticletitle{Flava: A foundational language and vision alignment model}. In \bibinfo{booktitle}{\emph{Proceedings of the IEEE/CVF Conference on Computer Vision and Pattern Recognition}}. \bibinfo{pages}{15638--15650}.
\newblock


\bibitem[Sun et~al\mbox{.}(2020)]%
        {sun2020conditional}
\bibfield{author}{\bibinfo{person}{Xin Sun}, \bibinfo{person}{Zhenning Yang}, \bibinfo{person}{Chi Zhang}, \bibinfo{person}{Keck-Voon Ling}, {and} \bibinfo{person}{Guohao Peng}.} \bibinfo{year}{2020}\natexlab{}.
\newblock \showarticletitle{Conditional gaussian distribution learning for open set recognition}. In \bibinfo{booktitle}{\emph{Proceedings of the IEEE/CVF conference on computer vision and pattern recognition}}. \bibinfo{pages}{13480--13489}.
\newblock


\bibitem[Wang et~al\mbox{.}(2025)]%
        {wang2025diffusion}
\bibfield{author}{\bibinfo{person}{Jinglong Wang}, \bibinfo{person}{Xiawei Li}, \bibinfo{person}{Jing Zhang}, \bibinfo{person}{Qingyuan Xu}, \bibinfo{person}{Qin Zhou}, \bibinfo{person}{Qian Yu}, \bibinfo{person}{Lu Sheng}, {and} \bibinfo{person}{Dong Xu}.} \bibinfo{year}{2025}\natexlab{}.
\newblock \showarticletitle{Diffusion model is secretly a training-free open vocabulary semantic segmenter}.
\newblock \bibinfo{journal}{\emph{IEEE Transactions on Image Processing}} (\bibinfo{year}{2025}).
\newblock


\bibitem[Wu et~al\mbox{.}(2023)]%
        {wu2023diffumask}
\bibfield{author}{\bibinfo{person}{Weijia Wu}, \bibinfo{person}{Yuzhong Zhao}, \bibinfo{person}{Mike~Zheng Shou}, \bibinfo{person}{Hong Zhou}, {and} \bibinfo{person}{Chunhua Shen}.} \bibinfo{year}{2023}\natexlab{}.
\newblock \showarticletitle{Diffumask: Synthesizing images with pixel-level annotations for semantic segmentation using diffusion models}.
\newblock \bibinfo{journal}{\emph{arXiv preprint arXiv:2303.11681}} (\bibinfo{year}{2023}).
\newblock


\bibitem[Wu et~al\mbox{.}(2019)]%
        {wu2019detectron2}
\bibfield{author}{\bibinfo{person}{Yuxin Wu}, \bibinfo{person}{Alexander Kirillov}, \bibinfo{person}{Francisco Massa}, \bibinfo{person}{Wan-Yen Lo}, {and} \bibinfo{person}{Ross Girshick}.} \bibinfo{year}{2019}\natexlab{}.
\newblock \bibinfo{title}{Detectron2}.
\newblock \bibinfo{howpublished}{\url{https://github.com/facebookresearch/detectron2}}.
\newblock


\bibitem[Xian et~al\mbox{.}(2019)]%
        {xian2019semantic}
\bibfield{author}{\bibinfo{person}{Yongqin Xian}, \bibinfo{person}{Subhabrata Choudhury}, \bibinfo{person}{Yang He}, \bibinfo{person}{Bernt Schiele}, {and} \bibinfo{person}{Zeynep Akata}.} \bibinfo{year}{2019}\natexlab{}.
\newblock \showarticletitle{Semantic projection network for zero-and few-label semantic segmentation}. In \bibinfo{booktitle}{\emph{Proceedings of the IEEE/CVF Conference on Computer Vision and Pattern Recognition}}. \bibinfo{pages}{8256--8265}.
\newblock


\bibitem[Xie et~al\mbox{.}(2023)]%
        {xie2023sed}
\bibfield{author}{\bibinfo{person}{Bin Xie}, \bibinfo{person}{Jiale Cao}, \bibinfo{person}{Jin Xie}, \bibinfo{person}{Fahad~Shahbaz Khan}, {and} \bibinfo{person}{Yanwei Pang}.} \bibinfo{year}{2023}\natexlab{}.
\newblock \showarticletitle{SED: A Simple Encoder-Decoder for Open-Vocabulary Semantic Segmentation}.
\newblock \bibinfo{journal}{\emph{arXiv preprint arXiv:2311.15537}} (\bibinfo{year}{2023}).
\newblock


\bibitem[Xu et~al\mbox{.}(2023a)]%
        {xu2023open}
\bibfield{author}{\bibinfo{person}{Jiarui Xu}, \bibinfo{person}{Sifei Liu}, \bibinfo{person}{Arash Vahdat}, \bibinfo{person}{Wonmin Byeon}, \bibinfo{person}{Xiaolong Wang}, {and} \bibinfo{person}{Shalini De~Mello}.} \bibinfo{year}{2023}\natexlab{a}.
\newblock \showarticletitle{Open-vocabulary panoptic segmentation with text-to-image diffusion models}. In \bibinfo{booktitle}{\emph{Proceedings of the IEEE/CVF Conference on Computer Vision and Pattern Recognition}}. \bibinfo{pages}{2955--2966}.
\newblock


\bibitem[Xu et~al\mbox{.}(2023b)]%
        {xu2023side}
\bibfield{author}{\bibinfo{person}{Mengde Xu}, \bibinfo{person}{Zheng Zhang}, \bibinfo{person}{Fangyun Wei}, \bibinfo{person}{Han Hu}, {and} \bibinfo{person}{Xiang Bai}.} \bibinfo{year}{2023}\natexlab{b}.
\newblock \showarticletitle{Side adapter network for open-vocabulary semantic segmentation}. In \bibinfo{booktitle}{\emph{Proceedings of the IEEE/CVF Conference on Computer Vision and Pattern Recognition}}. \bibinfo{pages}{2945--2954}.
\newblock


\bibitem[Xu et~al\mbox{.}(2022)]%
        {xu2022simple}
\bibfield{author}{\bibinfo{person}{Mengde Xu}, \bibinfo{person}{Zheng Zhang}, \bibinfo{person}{Fangyun Wei}, \bibinfo{person}{Yutong Lin}, \bibinfo{person}{Yue Cao}, \bibinfo{person}{Han Hu}, {and} \bibinfo{person}{Xiang Bai}.} \bibinfo{year}{2022}\natexlab{}.
\newblock \showarticletitle{A simple baseline for open-vocabulary semantic segmentation with pre-trained vision-language model}. In \bibinfo{booktitle}{\emph{European Conference on Computer Vision}}. Springer, \bibinfo{pages}{736--753}.
\newblock


\bibitem[Yang et~al\mbox{.}(2024)]%
        {yang2024generalized}
\bibfield{author}{\bibinfo{person}{Jingkang Yang}, \bibinfo{person}{Kaiyang Zhou}, \bibinfo{person}{Yixuan Li}, {and} \bibinfo{person}{Ziwei Liu}.} \bibinfo{year}{2024}\natexlab{}.
\newblock \showarticletitle{Generalized out-of-distribution detection: A survey}.
\newblock \bibinfo{journal}{\emph{International Journal of Computer Vision}} \bibinfo{volume}{132}, \bibinfo{number}{12} (\bibinfo{year}{2024}), \bibinfo{pages}{5635--5662}.
\newblock


\bibitem[Ye et~al\mbox{.}(2024)]%
        {ye2024differential}
\bibfield{author}{\bibinfo{person}{Tianzhu Ye}, \bibinfo{person}{Li Dong}, \bibinfo{person}{Yuqing Xia}, \bibinfo{person}{Yutao Sun}, \bibinfo{person}{Yi Zhu}, \bibinfo{person}{Gao Huang}, {and} \bibinfo{person}{Furu Wei}.} \bibinfo{year}{2024}\natexlab{}.
\newblock \showarticletitle{Differential transformer}.
\newblock \bibinfo{journal}{\emph{arXiv preprint arXiv:2410.05258}} (\bibinfo{year}{2024}).
\newblock


\bibitem[Yu et~al\mbox{.}(2023)]%
        {yu2023zero}
\bibfield{author}{\bibinfo{person}{Seonghoon Yu}, \bibinfo{person}{Paul~Hongsuck Seo}, {and} \bibinfo{person}{Jeany Son}.} \bibinfo{year}{2023}\natexlab{}.
\newblock \showarticletitle{Zero-shot Referring Image Segmentation with Global-Local Context Features}. In \bibinfo{booktitle}{\emph{Proceedings of the IEEE/CVF Conference on Computer Vision and Pattern Recognition}}. \bibinfo{pages}{19456--19465}.
\newblock


\bibitem[Zhang and Patel(2016)]%
        {zhang2016sparse}
\bibfield{author}{\bibinfo{person}{He Zhang} {and} \bibinfo{person}{Vishal~M Patel}.} \bibinfo{year}{2016}\natexlab{}.
\newblock \showarticletitle{Sparse representation-based open set recognition}.
\newblock \bibinfo{journal}{\emph{IEEE transactions on pattern analysis and machine intelligence}} \bibinfo{volume}{39}, \bibinfo{number}{8} (\bibinfo{year}{2016}), \bibinfo{pages}{1690--1696}.
\newblock


\bibitem[Zhao et~al\mbox{.}(2017)]%
        {zhao2017open}
\bibfield{author}{\bibinfo{person}{Hang Zhao}, \bibinfo{person}{Xavier Puig}, \bibinfo{person}{Bolei Zhou}, \bibinfo{person}{Sanja Fidler}, {and} \bibinfo{person}{Antonio Torralba}.} \bibinfo{year}{2017}\natexlab{}.
\newblock \showarticletitle{Open vocabulary scene parsing}. In \bibinfo{booktitle}{\emph{Proceedings of the IEEE International Conference on Computer Vision}}. \bibinfo{pages}{2002--2010}.
\newblock


\bibitem[Zhong et~al\mbox{.}(2021a)]%
        {ncd-ncl}
\bibfield{author}{\bibinfo{person}{Zhun Zhong}, \bibinfo{person}{Enrico Fini}, \bibinfo{person}{Subhankar Roy}, \bibinfo{person}{Zhiming Luo}, \bibinfo{person}{Elisa Ricci}, {and} \bibinfo{person}{Nicu Sebe}.} \bibinfo{year}{2021}\natexlab{a}.
\newblock \showarticletitle{Neighborhood contrastive learning for novel class discovery}. In \bibinfo{booktitle}{\emph{Proceedings of the IEEE/CVF Conference on Computer Vision and Pattern Recognition}}. \bibinfo{pages}{10867--10875}.
\newblock


\bibitem[Zhong et~al\mbox{.}(2021b)]%
        {openmix}
\bibfield{author}{\bibinfo{person}{Zhun Zhong}, \bibinfo{person}{Linchao Zhu}, \bibinfo{person}{Zhiming Luo}, \bibinfo{person}{Shaozi Li}, \bibinfo{person}{Yi Yang}, {and} \bibinfo{person}{Nicu Sebe}.} \bibinfo{year}{2021}\natexlab{b}.
\newblock \showarticletitle{Openmix: Reviving known knowledge for discovering novel visual categories in an open world}. In \bibinfo{booktitle}{\emph{Proceedings of the IEEE/CVF Conference on Computer Vision and Pattern Recognition}}. \bibinfo{pages}{9462--9470}.
\newblock


\bibitem[Zhou et~al\mbox{.}(2019)]%
        {zhou2019semantic}
\bibfield{author}{\bibinfo{person}{Bolei Zhou}, \bibinfo{person}{Hang Zhao}, \bibinfo{person}{Xavier Puig}, \bibinfo{person}{Tete Xiao}, \bibinfo{person}{Sanja Fidler}, \bibinfo{person}{Adela Barriuso}, {and} \bibinfo{person}{Antonio Torralba}.} \bibinfo{year}{2019}\natexlab{}.
\newblock \showarticletitle{Semantic understanding of scenes through the ade20k dataset}.
\newblock \bibinfo{journal}{\emph{International Journal of Computer Vision}}  \bibinfo{volume}{127} (\bibinfo{year}{2019}), \bibinfo{pages}{302--321}.
\newblock


\bibitem[Zhou et~al\mbox{.}(2022)]%
        {zhou2022extract}
\bibfield{author}{\bibinfo{person}{Chong Zhou}, \bibinfo{person}{Chen~Change Loy}, {and} \bibinfo{person}{Bo Dai}.} \bibinfo{year}{2022}\natexlab{}.
\newblock \showarticletitle{Extract free dense labels from clip}. In \bibinfo{booktitle}{\emph{European Conference on Computer Vision}}. Springer, \bibinfo{pages}{696--712}.
\newblock


\bibitem[Zhou et~al\mbox{.}(2023)]%
        {zhou2023zegclip}
\bibfield{author}{\bibinfo{person}{Ziqin Zhou}, \bibinfo{person}{Yinjie Lei}, \bibinfo{person}{Bowen Zhang}, \bibinfo{person}{Lingqiao Liu}, {and} \bibinfo{person}{Yifan Liu}.} \bibinfo{year}{2023}\natexlab{}.
\newblock \showarticletitle{Zegclip: Towards adapting clip for zero-shot semantic segmentation}. In \bibinfo{booktitle}{\emph{Proceedings of the IEEE/CVF Conference on Computer Vision and Pattern Recognition}}. \bibinfo{pages}{11175--11185}.
\newblock


\end{thebibliography}

\clearpage

\end{sloppypar}
\end{document}